\documentclass[journal]{IEEEtran}
\usepackage{url}
\usepackage[noadjust]{cite}
\usepackage{graphicx}
\usepackage{multirow}
\usepackage{algorithm}
\usepackage{algpseudocode}
\usepackage{comment}
\usepackage{makecell}
\usepackage{ragged2e}
\usepackage{rotating}
\usepackage{lscape}
\usepackage{nomencl}
\makenomenclature
\usepackage{wrapfig}
\usepackage{makecell}
\usepackage{balance}
\usepackage[utf8]{inputenc}
\usepackage{ragged2e}
\usepackage{booktabs}

\usepackage{pifont}
\usepackage{tabulary}
\usepackage{comment}
\usepackage{lipsum}
\usepackage{pifont}
\usepackage{graphicx}
\usepackage{array}
\usepackage{array, tabularx, multirow, booktabs, diagbox}
\usepackage{adjustbox}
\usepackage{amsthm}

\theoremstyle{remark}

\usepackage[table,xcdraw]{xcolor} 
\usepackage{colortbl} 
\usepackage{amssymb} 

\newcommand{\cmark}{\ding{51}}%
\newcommand{\xmark}{\ding{55}}%
\usepackage{amsmath,amssymb,amsfonts}
\usepackage{graphicx}
\usepackage{textcomp}
\usepackage{subcaption}
\usepackage[font={small}]{caption}
\usepackage{epstopdf}
\usepackage{pdfpages}
\usepackage{algorithmicx}
\usepackage{epsfig}
\usepackage{amsmath}
\usepackage{amssymb}
\usepackage{algorithm}
\usepackage{pifont}
\usepackage{bm}
\usepackage{authblk}
\usepackage[utf8]{inputenc}
\usepackage{tabularx}
\usepackage[inline]{enumitem}
\usepackage{multirow}
\usepackage{commath}
\UseRawInputEncoding
\begin{document}
\title{Vision-Language Models for Edge Networks: A Comprehensive Survey}

\author{Ahmed~Sharshar,~Latif~U.~Khan,~\IEEEmembership{Member,~IEEE},~Waseem~Ullah,~\IEEEmembership{Member,~IEEE},~Mohsen~Guizani,~\IEEEmembership{Fellow,~IEEE}

        
    
\IEEEcompsocitemizethanks{
 \IEEEcompsocthanksitem Corresponding author: Ahmed Sharshar {ahmed.sharshar@mbzuai.ac.ae}
\IEEEcompsocthanksitem A. Sharshar is affiliated with the Computer Vision Department, while L. U. Khan, W. Ullah, and M. Guizani are with the Machine Learning Department, all at Mohamed Bin Zayed University of Artificial Intelligence (MBZUAI), Abu Dhabi, UAE.

}}

\markboth{}{}%

\maketitle





\begin{abstract} 
Vision Large Language Models (VLMs) combine visual understanding with natural language processing, enabling tasks like image captioning, visual question answering, and video analysis. While VLMs show impressive capabilities across domains such as autonomous vehicles, smart surveillance, and healthcare, their deployment on resource-constrained edge devices remains challenging due to processing power, memory, and energy limitations. This survey explores recent advancements in optimizing VLMs for edge environments, focusing on model compression techniques, including pruning, quantization, knowledge distillation, and specialized hardware solutions that enhance efficiency. We provide a detailed discussion of efficient training and fine-tuning methods, edge deployment challenges, and privacy considerations. Additionally, we discuss the diverse applications of lightweight VLMs across healthcare, environmental monitoring, and autonomous systems, illustrating their growing impact. By highlighting key design strategies, current challenges, and offering recommendations for future directions, this survey aims to inspire further research into the practical deployment of VLMs, ultimately making advanced AI accessible in resource-limited settings.
\end{abstract}


\section{Introduction}

\begin{figure*}[t]
    \centering
    \includegraphics[width=\textwidth]{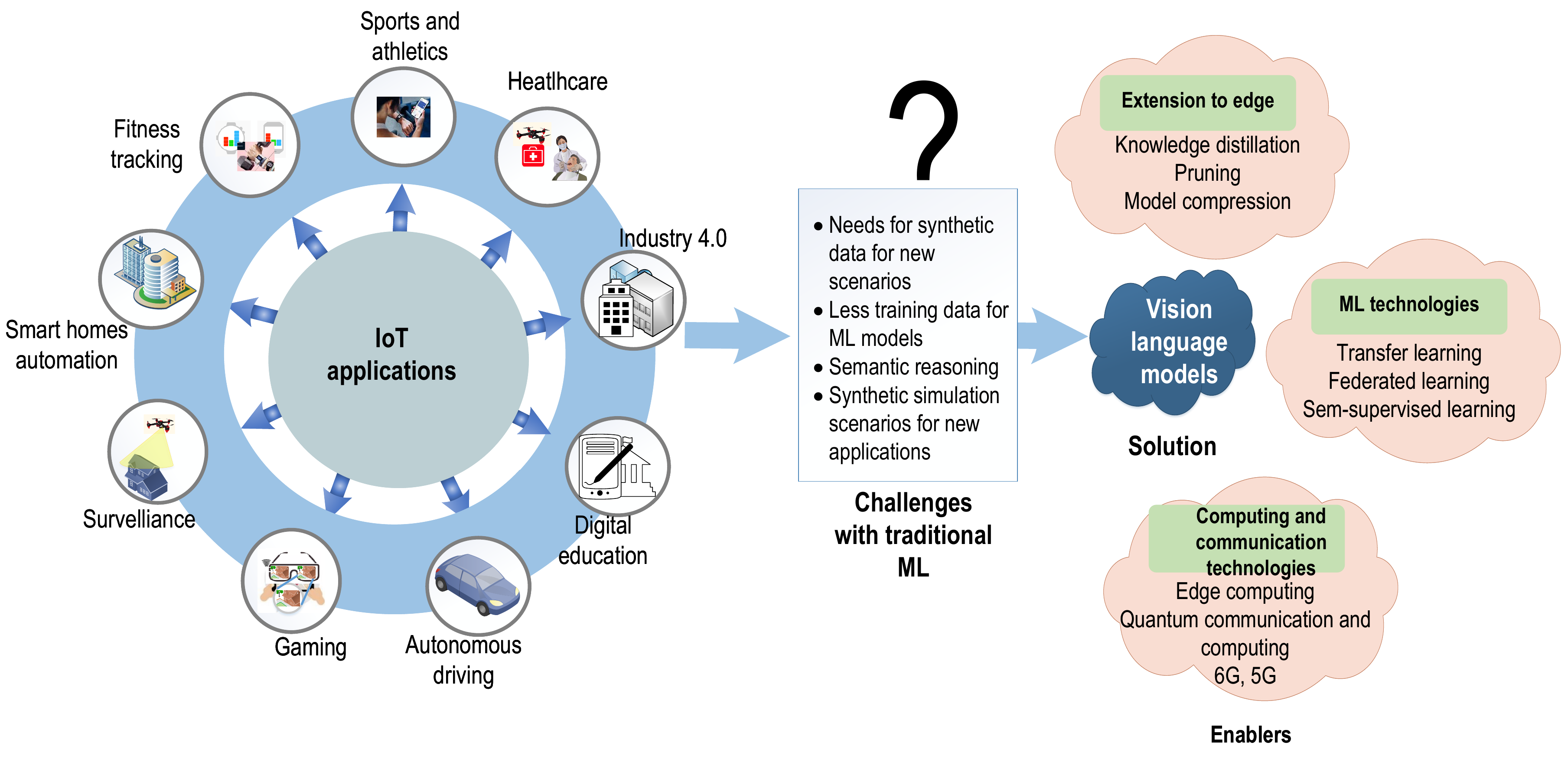}
    \caption{\textcolor{black}{Overview of integrating Vision-Language Models (VLMs) into edge networks. The diagram highlights IoT applications (e.g., healthcare, autonomous driving, smart homes, surveillance, gaming, education, fitness tracking, Industry 4.0, sports) that benefit from VLMs to address traditional Machine Learning (ML) challenges such as limited training data and semantic reasoning. It outlines key ML techniques (transfer learning, federated learning, semi-supervised learning) and model optimization methods (knowledge distillation, pruning, compression) for efficient deployment. Crucial enabling technologies, including edge computing, quantum communication and computing, and advanced wireless networks (5G/6G), are also presented to support VLM deployment at the edge}
    \label{fig:Main Overview}}
\end{figure*}

The integration of vision and language understanding in artificial intelligence has given rise to VLMs, which combine visual inputs with natural language processing to perform tasks such as image captioning, visual question answering, and visual content generation \cite{Lu2019, Li2020, Kim2021,khan2024edge}. These models have demonstrated promising capabilities in various domains, from social media content moderation to assisting autonomous vehicle navigation, enabling machines to interact with their environment more intuitively and human-likely. Although VLMs offer many benefits, it is challenging to extend VLMs at the network edge. Extending VLMs to edge devices remains very challenging due to resource limitations of edge devices (e.g., smartphone and wearable). Edge devices, characterized by their limited processing power, memory, and energy consumption, require VLMs that are accurate but also lightweight and efficient \cite{Li2019, Wu2020}. The challenges posed by these constraints necessitate innovative approaches to model design and optimization to ensure that VLMs can be effectively deployed on edge platforms \cite{lin2023pushing}.

Recent studies have aimed to compress VLMs and use edge deployment with pruning, quantization, and Knowledge Distillation methods \cite{efficientvlm2024}. Pruning consists of removing redundant or insignificant parameters from the model, reducing the model's size and computational overhead while maintaining similar performance \cite{Han2016}. Quantization reduces the precision of model weights and activations, which can greatly impact memory usage and inference speed \cite{Jacob2018}. In knowledge distillation, knowledge from a large, cumbersome model (teacher) is distilled into a smaller model (student) \cite{Hinton2015}. Moreover, purpose-built hardware accelerators (e.g., Google’s Edge TPU) and edge-native architectures have also played a crucial role in enhancing the accessibility in deploying VLMs on edge-constrained hardware \cite{Sze2017, Reddi2020}. This survey highlights these advancements and presents a comprehensive overview of lightweight visual language models (VLMs) for edge applications, discussing the trade-offs involved in striking the balance between model efficiency and performance.

\subsection{Motivation}
The demand for real-time processing of visual tasks, for example, autonomous driving, smart surveillance, and augmented reality, is one of the primary reasons to deploy VLMs on edge devices \cite{Chen2019, Wang2020, Tan2021}. ITS, one of the crucial applications, includes object detection, traffic sign recognition, and pedestrian detection. Offloading this processing to the cloud incurs latency, impeding time-critical use cases. Likewise, edge processing in smart surveillance: Processing video features on edge devices (e.g., IP cameras) protects the privacy of target information by reducing private data transmission over networks \cite{Lane2015}. Augmented reality applications also use low-latency processing to interact seamlessly. These applications are made possible by lightweight VLMs that guarantee high-performance resource usage at the edge \cite{Jouppi2017}. Calculations performed closer to the data reduce latency and add reliability by reducing reliance on stable connections.

The use of VLMs on edge devices faces challenges. Current VLMs are unusually large and do not fit into the memory/storage of most edge devices. For instance, the GPT-3 model with 175 billion parameters demands about 350 GB for inference memory alone \cite{Brown2020}, while CLIP, a common VLM, has 63 million parameters \cite{Radford2021}, which is not appropriate for edge devices and other limited resources regions. These models require considerable hardware resources and energy, which is typically a limitation for low backup energy devices. Edge devices, including smartphones and Internet of Things (IoT) sensors, are typically equipped with a backup energy capacity of 1,000 mAh to 5,000 mAh \cite{Zanella2020}, making it challenging to run power-hungry computations locally with these models.

In addition, when such models perform inference, they may quickly consume the energy supplies of edge devices, restricting their operational time and efficiency \cite{Zhang2020}. Moreover, the computational complexity of these models often requires hardware accelerators (GPUs, TPUs, etc.), which may not be realistic for many edge computing applications due to cost and power availability \cite{Chen2021}. Figure \ref{fig:Main Overview} shows an overview of using VLM in the IOT framework.

We address these issues with a balance of model complexities over resource allocations. Compromises in performance and accuracy are often required to balance model complexity against resource efficiency in addressing these issues. One specific type of low-resource VLMs is model compression methods, including pruning, quantization, and knowledge distillation, which can significantly decrease VLMs' size and computational cost. However, this may result in a momentary drop in accuracy, which ought to be judiciously managed to make the model effective for the application \cite{Han2016, Jacob2018, Hinton2015}.

The second key issue comes from the devices' heterogeneity in computing power and local energy reserves. This variability adds complexity to the deployment process for VLMs, as the models must be tuned to the capabilities of each device. Some works, such as dynamic inference, have attempted this by scaling the network with respect to the computation available at inference, balancing resource use and accuracy \cite{Tan2019, Yu2019}. Model scaling, for instance, makes several models of different complexities so they can be deployed on edge devices with differing capabilities. Dynamic inference techniques can account for this by adjusting the computation at inference time and balancing speed and accuracy based on real-time resource availability.

Further research is, however, required to create solutions that are domain-agnostic and can be adapted to the different constraints of different edge environments. To facilitate edge-native VLM adoption, it is crucial to have these models work with reasonable efficiency over significant device variance with minor device specialization.

Furthermore, optimizing VLMs for edge devices includes research on novel model architectures with lower computational requirements by design. One such adaptation, for example, is using transformer-based models that offer more efficient variants for edge deployment \cite{Vaswani2017, Chen2020}. These adaptations generally include simplifying the attention mechanisms or reducing the layers in the model to reduce computational overhead. In addition, there is a trend of using attention approaches and lightweight CNNs to achieve trade-offs between effectiveness and resources \cite{Hu2018, Howard2019}. To boost performance on mobile devices, MobileNetV3 is proposed with architectural innovations: depthwise separable convolutions that compress the number of parameters and the computations required \cite{Howard2019}. These architectural advances play an important role in expanding the potential of what is possible with VLMs on edge devices, allowing more capable models to run within the constraints of simpler hardware.

Lightweight VLMs can have a variety of application domains that are growing rapidly. For instance, through VLMs, medical image analysis and diagnostics can be performed directly on portable devices, enabling immediate feedback and decision support \cite{Esteva2017, Topol2019}. This capability is invaluable in remote or resource-poor environments where access to advanced medical care is restricted. Through visual recognition and language understanding capabilities, VLMs allow for next-generation inventory management and customer interaction in retail \cite{Ren2015, Liu2016}. For example, VLMs can examine smart shop assistants that identify and explain products in detail to customers seamlessly and grammatically. These models can have implications for multiple domains, showcasing the potential versatility of lightweight VLMs. VLMs cover a tremendous spectrum of applications, from driving efficiency in industries where real-time visual inspections can be automated through VLMs to enabling everyday experiences for those with disabilities by describing the contents of the camera stream captured in the real world.

\begin{table*}[ht]
\centering
\caption{Comparison of Various Studies on Vision-Language Models (VLMs).}
\renewcommand{\arraystretch}{1.2} 
\rowcolors{1}{gray!15}{white} 
\begin{tabular}{l c c c c p{4.5cm}} 
\hline
\textbf{Reference} & \textbf{Security and Privacy} & \textbf{Efficient Fine Tuning} & \textbf{On Edge Inference} & \textbf{Applications} & \textbf{Remark} \\ \hline
Du \emph{et al.} \cite{surveyvisionlanguagepretrainedmodels} & \xmark & \cmark & \xmark & \cmark & This work surveys vision-language pre-trained models, focusing on their architectures, training methods, and applications. \\  \hline

Li \emph{et al.} \cite{visionlanguageintelligencetasksrepresentation} & \xmark & \cmark & \xmark & \xmark & The study explores vision-language intelligence, emphasizing tasks, representation learning, and the development of large models. \\ \hline

Xing \emph{et al.} \cite{SurveyofEfficientFine-TuningMethods} & \cmark & \cmark & \xmark & \xmark & The paper provides an overview of efficient fine-tuning methods for vision-language models, with a focus on Prompt and Adapter techniques. \\ \hline

Ghosh \emph{et al.} \cite{ghosh2024exploringfrontiervisionlanguagemodels} & \xmark & \xmark & \xmark & \cmark & This article reviews current methodologies and future directions of vision-language models, with emphasis on their development and applications. \\ \hline

Zhang \emph{et al.} \cite{zhang2024visionlanguagemodelsvisiontasks} & \xmark & \cmark & \xmark & \cmark & This study highlights vision-language models for vision tasks, emphasizing their theoretical foundations and practical applications. \\ \hline

Cui \emph{et al.} \cite{ASurveyonMultimodalforAutonomousDrivin} & \xmark & \cmark & \xmark & \cmark & This review discusses multimodal large language models for autonomous driving, with attention to their applications and efficiency. \\ \hline

Yin \emph{et al.} \cite{surveymultimodallargelanguage} & \cmark & \xmark & \xmark & \cmark & This comprehensive study offers an in-depth overview of multimodal large language models. \\ \hline

Jin \emph{et al.} \cite{efficientmultimodallargelanguage} & \xmark & \cmark & \xmark & \cmark & This research examines efficient multimodal large language models, focusing on their design and applications. \\ \hline

Our Survey & \cmark & \cmark & \cmark & \cmark & N/A\\
\hline
\end{tabular}%
\label{tab:literaturecomparison}
\end{table*}

\subsection{Market Statistics and Research Trends}
VLMs have rapidly emerged as a new market, driven by demand for systems that understand and reason with visual and textual information. The global AI market was valued at USD 58.3 billion in 2021 and is projected to reach USD 309.6 billion by 2026, with a CAGR of 39.7\% \cite{MarketsandMarkets2021}. The VLM segment is expected to grow at the fastest rate. Its market size is estimated to reach \$2.5 billion in 2024, up from \$1.8 billion in 2023 and \$1.2 billion in 2022 \cite{huggingface2024}. VLMs are gaining traction across sectors such as healthcare, automotive, and consumer electronics. In the automotive industry, VLMs contribute to ADAS and autonomous driving solutions. With the rise of smart devices and IoT, there is increasing demand for lightweight VLMs capable of running efficiently on edge devices.

Research trends in VLMs emphasize improving efficiency and accuracy while minimizing computational cost. Studies have explored model compression techniques such as pruning, quantization, and knowledge distillation \cite{Han2016, Hinton2015, Jacob2018}. There is also growing interest in hybrid architectures combining CNNs and transformers \cite{Vaswani2017, Howard2019}, aiming to merge CNNs’ efficiency with transformers’ representational power. Multi-task learning is another trend, where a single VLM handles multiple tasks, enhancing efficiency and reducing reliance on task-specific models. The number of research papers on VLMs and AI for edge devices has notably increased, reflecting heightened academic interest.

VLM applications are expanding rapidly, with major investments in healthcare, retail, and security. In healthcare, they support medical image analysis, diagnosis, and telemedicine \cite{Esteva2017, Topol2019}. Retail uses include smart shopping assistants and personalized marketing \cite{Ren2015, Liu2016}. In security, automated surveillance leverages VLMs to detect anomalies and threats. These examples show VLMs' versatility and growing impact. Edge AI, crucial for real-time processing and privacy, deploys models on devices like smartphones and IoT sensors, bypassing cloud dependency \cite{Lane2015}. This shift is motivated by latency, bandwidth, and privacy issues. Research in edge AI focuses on model optimization and hardware accelerators for efficient inference on constrained devices \cite{Sze2017, Reddi2020}. Companies like NVIDIA, Intel, and Google are investing heavily in this area. The global edge AI hardware market is projected to reach USD 3.89 billion by 2025, growing at a CAGR of 20.6\% from 2018 \cite{AlliedMarketResearch2020}.

\subsection{Existing Surveys and Tutorials}
Few surveys have reviewed VLMs, their efficiency, and applications \cite{surveyvisionlanguagepretrainedmodels, visionlanguageintelligencetasksrepresentation, SurveyofEfficientFine-TuningMethods, ghosh2024exploringfrontiervisionlanguagemodels, zhang2024visionlanguagemodelsvisiontasks, ASurveyonMultimodalforAutonomousDrivin, surveymultimodallargelanguage, efficientmultimodallargelanguage}. Table \ref{tab:literaturecomparison} summarizes key scopes and how they differ from ours.

\cite{surveyvisionlanguagepretrainedmodels} focused on vision-language pre-trained models, covering their evolution, architectures, and integration methods. \cite{visionlanguageintelligencetasksrepresentation} explored vision-language intelligence, emphasizing tasks, representation learning, and large model development, along with performance insights and research directions. Xing et al. \cite{SurveyofEfficientFine-TuningMethods} surveyed efficient fine-tuning methods for VLMs, focusing on Prompt and Adapter techniques and related challenges. Ghosh et al. \cite{ghosh2024exploringfrontiervisionlanguagemodels} provided an overview of current methodologies and future directions, highlighting strengths, limitations, and areas for exploration. Zhang et al. \cite{zhang2024visionlanguagemodelsvisiontasks} discussed VLMs for vision tasks, covering theoretical foundations, practical applications, and challenges in domains like medical imaging and industrial automation. \cite{ASurveyonMultimodalforAutonomousDrivin} focused on multimodal models for autonomous driving, discussing modality integration, performance methods, and specific applications. Yin et al. \cite{surveymultimodallargelanguage} reviewed multimodal large language models with emphasis on efficient design, architectures, and applications in biomedical analysis and document understanding. Finally, Jin et al. \cite{efficientmultimodallargelanguage} surveyed efficient multimodal LLMs, focusing on reducing computational cost, improving efficiency, and applications in high-res image understanding and medical QA, while outlining challenges and research directions.

Different from existing works \cite{surveyvisionlanguagepretrainedmodels, visionlanguageintelligencetasksrepresentation, SurveyofEfficientFine-TuningMethods, ghosh2024exploringfrontiervisionlanguagemodels, zhang2024visionlanguagemodelsvisiontasks, ASurveyonMultimodalforAutonomousDrivin, surveymultimodallargelanguage, efficientmultimodallargelanguage}, we present a comprehensive overview of VLMs, including key design aspects and high-level architecture. We also provide deployment challenges on edge devices. Furthermore, several open research challenges are discussed, along with promising solution approaches.

\subsection{Our Survey}
This survey aims to examine the techniques, architectures, and applications that define the rapidly evolving area of VLMs for edge networks. By addressing the challenges and showcasing the solutions, this paper contributes to the ongoing efforts to make sophisticated VLMs accessible and practical for edge computing environments. The continued innovation in this field promises to unlock new capabilities and applications, bringing the power of AI-driven vision and language understanding to a broader range of devices and use cases. Our survey aims to answer the following questions:
\begin{itemize}
    \item How do we efficiently enable VLM at the network edge? 
    \item What are the existing schemes and their limitations that will help deploy VLM at the network edge?
    \item How does one enable secure and privacy-ware VLM?
    \item What are the challenges and their possible solutions in allowing VLMs to at the network edge?
    \item What are the different application domains for VLMs, and what opportunities are available?
\end{itemize} 

Our contributions are summarized as follows:
\begin{itemize}
    \item We present the key concepts, main design aspects, and high-level architecture for Vision-Language Models.
    \item A comprehensive cycle for extending the VLMs from the cloud to the edge is provided, considering efficient training and fine-tuning methods, edge deployment challenges, and privacy and security issues. We consider issues related to designing efficient VLMs, deploying them on edge devices, addressing privacy and security concerns, and enhancing their performance on low-resource devices.
    \item Several open challenges are presented, including the difficulties of deploying VLMs on edge devices and fine-tuning them with limited resources. Moreover, we discussed about promising solution approaches.
\end{itemize}

\section{Fundamentals of Vision Language Models}

\begin{figure*}[htbp]
    \centering
    \begin{minipage}{0.48\textwidth}
        \centering
        \includegraphics[width=0.95\textwidth]{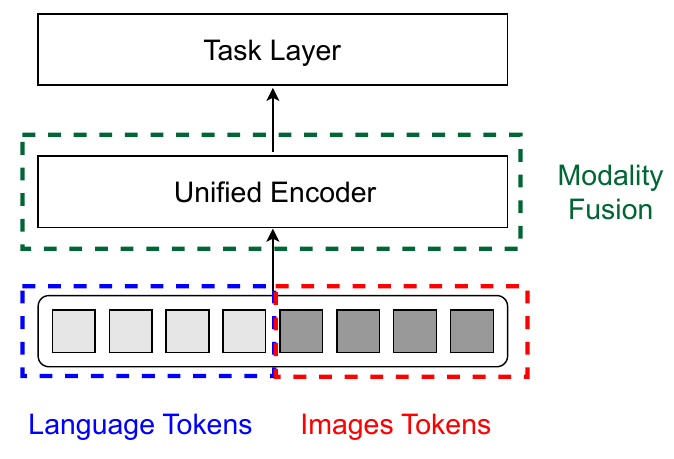}
        \caption*{Single Stream}
    \end{minipage}
    \hfill
    \begin{minipage}{0.48\textwidth}
        \centering
        \includegraphics[width=0.95\textwidth]{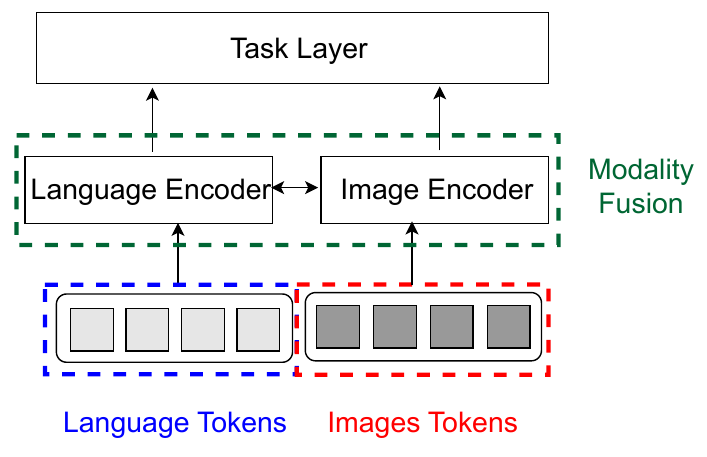}
        \caption*{Dual Stream}
    \end{minipage}
    \caption{The architecture of Vision-Language Pre-trained (VLP) models typically includes three key components: Visual Embedding (VE), Textual Embedding (TE), and Modality Fusion (MF). Fig. (a) illustrates a dual-stream model, while Fig. (b) depicts a single-stream model. In dual-stream models, modality fusion can be optional and generally occurs through interactions (often via cross-attention) between the separate language and image encoders. Conversely, in single-stream models, modality fusion is inherently integrated within a unified encoder, which is typically a multi-layer transformer \cite{visionlanguageintelligencetasksrepresentation}.}
    \label{Single vs dual}
\end{figure*}

VLMs are designed to process and integrate visual and textual information simultaneously. These models leverage the combined power of computer vision and natural language processing to perform various multimodal tasks such as image captioning, visual question answering (VQA), and image-text retrieval. This section provides a detailed theoretical understanding of how VLMs work, including their mathematical representation and model architectures.

\subsection{Key Concepts}
They learn to align visual and textual modalities in a shared representation space, enabling cross-modal understanding and interaction. This process is a series of steps per modality (text and image) of tokenization, embedding, and encoding. In doing so, VLMs are able to model rich semantic interactions both within a single modality and cross-modality as one unified feature that connects image, text, and sound representations, improving downstream tasks like captioning, retrieval, and question answering.

\textbf{Text Representation}

Assuming a text input sequence 
\[
T = [t_1, t_2, \ldots, t_N],
\]
where each \( t_i \) is the \( i \)-th token (e.g., an integer ID or a subword unit). We first map each token \(t_i\) into a high-dimensional word embedding \(\mathbf{e}_i \in \mathbb{R}^{d}\). Concretely, one may define a learnable embedding matrix \(\mathbf{W}_{E} \in \mathbb{R}^{d \times |\mathcal{V}|}\) (where \(|\mathcal{V}|\) is the vocabulary size), so that each \(t_i\) (represented as a one-hot or index) is projected into \(\mathbf{e}_i\):
\textcolor{black}{
\begin{equation}
\mathbf{e}_i = \mathbf{W}_{E} \cdot \mathrm{one\_hot}(t_i) + \mathbf{b}_{E},
\label{eqn:word_embed}
\end{equation}}
where \(\mathbf{b}_{E}\in \mathbb{R}^d\) is a bias term and \(\mathrm{one\_hot}(t_i)\in\mathbb{R}^{|\mathcal{V}|}\) is a sparse vector with 1 at position \(t_i\) and 0 elsewhere. Equivalently, modern frameworks often bypass the explicit one-hot representation by using an embedding lookup.

In order to account for the sequence information, we add a positional embedding \(\mathbf{p}_i \in \mathbb{R}^{d}\) to each token embedding. Often, \(\mathbf{p}_i\) is either a learned parameter or derived from a fixed function (e.g., sinusoidal) \cite{vaswani2017attention}. The total embedding \(\mathbf{h}_i\) of the \(i\)-th token is then:
\textcolor{black}{
\begin{equation}
\mathbf{h}_i = \mathbf{e}_i + \mathbf{p}_i,
\label{eqn:pos_embed}
\end{equation}}\\
\textcolor{black}{where $\mathbf{h}_i$ denotes the final representation of the $i^{th}$ token. $\mathbf{e}_i$ denotes the embedding vector. Moreover, $\mathbf{p}_i$ represents the positional encoding}. These combined embeddings \(\{\mathbf{h}_i\}_{i=1}^{N}\) are subsequently processed through multiple transformer layers or other neural encoders to build contextualized text representations. For example, denoting one transformer block by \(\mathrm{TransBlock}(\cdot)\), a simplified representation of one layer’s output \(\mathbf{h}_i^{(l)}\) (for the \(l\)-th layer) could be:
\textcolor{black}{
\begin{equation}
\mathbf{h}_i^{(l)} = \mathrm{TransBlock}\bigl(\{\mathbf{h}_1^{(l-1)}, \ldots, \mathbf{h}_N^{(l-1)}\}\bigr)_i,
\label{eqn:transformer_layer}
\end{equation}}
where \(\mathbf{h}_i^{(0)} = \mathbf{h}_i\). Stacking multiple layers refines the representation of each token based on its context in the sequence. Ultimately, one can use \(\{\mathbf{h}_i^{(L)}\}\) from the final layer \(L\) for downstream tasks or for multimodal fusion.

\textbf{Image Representation}

For the image input \( I \), we extract informative visual features that reflect the image’s semantic and spatial content. Depending on the model’s backbone, either convolutional neural networks (CNNs) or Vision Transformers (ViTs) can be used. Let us divide the image \( I \) into patches \(\{ I_j \}_{j=1}^{M}\), where each patch covers a local region of the image. A learnable vision encoder is then applied to each patch:
\textcolor{black}{
\begin{equation}
\mathbf{v}_j = \mathrm{VisionEncoder}(I_j),
\label{eqn:vision_patch}
\end{equation}}
with \(\mathbf{v}_j\in\mathbb{R}^d\). In CNN-based backbones, \(\mathrm{VisionEncoder}(\cdot)\) might represent the final convolutional layers, extracting feature maps that encode both low-level and high-level cues. In Vision Transformers \cite{dosovitskiy2020image}, one often computes a linear projection for each patch, adds learnable 2D positional embeddings, and processes them through multi-head self-attention blocks, analogous to the transformer encoder in text processing. Concretely, for a ViT, one may write
\[
\mathbf{v}_j^{(0)} = \mathrm{PatchEmbed}(I_j) + \mathbf{p}_j^\text{(img)},
\]
followed by repeated transformer layers:
\textcolor{black}{
\begin{equation}
\mathbf{v}_j^{(l)} = \mathrm{TransBlock}\bigl(\{\mathbf{v}_1^{(l-1)}, \ldots, \mathbf{v}_M^{(l-1)}\}\bigr)_{j}, 
\end{equation}}
to yield the final visual tokens \(\{\mathbf{v}_j^{(L)}\}\).

The result \(\{\mathbf{v}_j\}_{j=1}^{M}\) (or the final-layer embeddings in the case of a ViT) provides a set of patch-level representations that can be mapped into the same dimensional space as \(\mathbf{h}_i\). By aligning \(\mathbf{v}_j\) and \(\mathbf{h}_i\) in a unified embedding space, the model can learn cross-modal correspondences.

\subsection{Mechanisms for Vision-Language Interaction}

At the center of VLMs is the integration of textual and visual embeddings. There are two main architectures to achieve this — fusion and dual encoders. Fig. \ref{Single vs dual} illustrates the key dissimilarity between the two Architectures.

\textbf{Single-Stream Architecture (Fusion Encoders):}

In contrast, single-stream models do early fusion by interleaving visual and textual encodings into a single sequence fed through a common encoder — often, a transformer \cite{li2019visualbert,su2019vlbert}. This architecture relies on the assumption that a single transformer encoder can adequately model the interactions among the modalities. Concretely, denote the final text embeddings by \(\{\mathbf{h}_i^{(L_t)}\}_{i=1}^{N}\) and the final image embeddings by \(\{\mathbf{v}_j^{(L_v)}\}_{j=1}^{M}\). These are concatenated into a unified sequence:
\[
[\mathbf{h}_1^{(L_t)}, \mathbf{h}_2^{(L_t)}, \ldots, \mathbf{h}_N^{(L_t)}; \ \mathbf{v}_1^{(L_v)}, \mathbf{v}_2^{(L_v)}, \ldots, \mathbf{v}_M^{(L_v)}].
\]
A shared multimodal transformer, \(\mathrm{Transformer}(\cdot)\), is then applied to capture deep interactions between the textual and visual tokens. Formally, the output embedding \(\mathbf{z}_k\) at position \(k\) in the fused sequence is:
\textcolor{black}{
\begin{equation}
\mathbf{z}_k = \mathrm{Transformer}\Bigl(\bigl[\mathbf{h}_1^{(L_t)}, \ldots, \mathbf{h}_N^{(L_t)}; \ \mathbf{v}_1^{(L_v)}, \ldots, \mathbf{v}_M^{(L_v)}\bigr]\Bigr)_k.
\label{eqn:fusion_transformer}
\end{equation}}
This single-stream approach encodes both modalities together, allowing each token (text or patch) to directly attend to and influence every other token. Such early fusion can be advantageous for tasks demanding fine-grained alignment between language and vision, such as Visual Question Answering and grounded dialogue systems.

A single model that can perform all tasks is usually a huge benefit because the implementation is much simpler and more efficient. They reduce memory and potential inference times by using one encoder instead of two, simplifying the architecture. Moreover, such a unified approach becomes a powerful tool for tasks demanding rich interaction between text and image, like image captioning and VQA. This has been evidenced by models like ViLT \cite{kim2021vilt}, which utilize a vision-and-language transformer without convolutional or region-based supervision and still perform strongly.

However, the single-stream approach has its problems, too, such as the increasing computational burden as longer sequences have to be concatenated and processed, which can be computationally intensive. In addition, the model has to learn from both modalities simultaneously, resulting in potentially non-ideal performance.

\begin{table*}[ht]
\centering
\caption{\textcolor{black}{A Comparison Between Some Lightweight Vision-Language Models.}}
\renewcommand{\arraystretch}{1.5}
\begin{tabular}{l c c p{2cm} p{5cm}}
\hline
\textcolor{black}{\textbf{Model}} & \textcolor{black}{\textbf{Year}} & \textcolor{black}{\textbf{Fusion Scheme}} & \textcolor{black}{\textbf{Parameters}} & \textcolor{black}{\textbf{Applications}} \\ \hline

\textcolor{black}{MobileVLM V2 \cite{mobilevlmv2}} & \textcolor{black}{2024} & \textcolor{black}{Single stream} & \textcolor{black}{1.1B} & \textcolor{black}{Mobile applications, Real-time image captioning} \\  \hline

\textcolor{black}{EfficientVLM \cite{efficientvlm2024}} & \textcolor{black}{2024} & \textcolor{black}{Single stream} & \textcolor{black}{92M} & \textcolor{black}{Visual question answering, Image retrieval} \\  \hline

\textcolor{black}{Unified-IO \cite{unifiedio2024}} & \textcolor{black}{2024} & \textcolor{black}{Single stream} & \textcolor{black}{1B} & \textcolor{black}{Integrated multimodal tasks, Visual question answering} \\  \hline

\textcolor{black}{LightVLP \cite{lightvlp2024}} & \textcolor{black}{2024} & \textcolor{black}{Dual stream} & \textcolor{black}{Not specified} & \textcolor{black}{Cross-modal retrieval, Visual grounding} \\  \hline

\textcolor{black}{Xmodel-VLM \cite{xmodelvlm2024}} & \textcolor{black}{2024} & \textcolor{black}{Single stream} & \textcolor{black}{1.1B} & \textcolor{black}{Text-image alignment, Visual question answering} \\  \hline

\textcolor{black}{EM-VLM4AD \cite{emvlm4ad2024}} & \textcolor{black}{2024} & \textcolor{black}{Single stream} & \textcolor{black}{223M (T5-Base) / 750M (T5-Large)} & \textcolor{black}{Autonomous driving, Traffic behavior prediction} \\  \hline

\textcolor{black}{ViTamin \cite{Chen_2024_CVPR}} & \textcolor{black}{2024} & \textcolor{black}{Single stream} & \textcolor{black}{Not specified} & \textcolor{black}{Image classification, Open-vocabulary detection} \\  \hline

\textcolor{black}{LINGO-2 \cite{wayve2024lingo}} & \textcolor{black}{2024} & \textcolor{black}{Single stream} & \textcolor{black}{5B} & \textcolor{black}{Autonomous driving, Driving behavior prediction} \\  \hline

\textcolor{black}{InstructBLIP \cite{instructblip2024}} & \textcolor{black}{2024} & \textcolor{black}{Single stream} & \textcolor{black}{Not specified} & \textcolor{black}{Instruction tuning, Question answering, Image captioning} \\  \hline

\textcolor{black}{RAVEN \cite{raven2024}} & \textcolor{black}{2024} & \textcolor{black}{Single stream} & \textcolor{black}{Not specified} & \textcolor{black}{Visual question answering, Image captioning} \\  \hline

\textcolor{black}{ScreenAI \cite{screenai2024}} & \textcolor{black}{2024} & \textcolor{black}{Single stream} & \textcolor{black}{Not specified} & \textcolor{black}{UI understanding, Infographic analysis} \\  \hline

\textcolor{black}{ALLaVA \cite{allava2024}} & \textcolor{black}{2022} & \textcolor{black}{Single stream} & \textcolor{black}{Not specified} & \textcolor{black}{Vision-language instruction tuning, Data synthesis} \\  \hline

\textcolor{black}{MiniVLM \cite{minivlm2022}} & \textcolor{black}{2022} & \textcolor{black}{Single stream} & \textcolor{black}{45.7M} & \textcolor{black}{Lightweight image-text processing, Visual question answering} \\  \hline

\end{tabular}
\label{tab:comparison}
\end{table*}

\textbf{Dual-Stream Architecture (i.e., Dual Encoders):}
On the other hand, dual-stream models adopt independent encoders for both visual and textual data, encode each modality separately, and then join their representation either through cross-attention mechanisms or other approaches. This architecture is especially useful when each input modality has limited overlapping features and can be processed differently. These independent processing streams are then merged in a higher-level step (usually through a cross-modal attention mechanism) that allows the model to learn how the modalities interact with each other after being processed and encoded independently. Treating individual modality streams with flexible structures provides full flexibility and may lead to more robust performance, as the model can capture and preserve the unique characteristics of each modality before combining them. Text and image embeddings are processed independently and later merged in the dual-stream architecture \cite{radford2021learning,lu2019vilbert,tan2019lxmert}:
\begin{align}
    \mathbf{h}'_i &= \text{TextEncoder}(\mathbf{h}_i), \\
    \mathbf{v}'_j &= \text{ImageEncoder}(\mathbf{v}_j), \\
    \mathbf{z}_k &= \text{CrossAttention}([\mathbf{h}'_1, \ldots, \mathbf{h}'_N], [\mathbf{v}'_1, \ldots, \mathbf{v}'_M]).
\end{align}

The most notable are dual-stream models, such as ViLBERT \cite{lu2019vilbert} and LXMERT \cite{tan2019lxmert}, which use separate transformers for image and text. This is especially useful for tasks in which the relationships between the modalities are complex and need to be modeled in detail, such as VQA and image-text retrieval. Because each stream can process and encode its own domain separately, dual-stream models may outperform single-stream models on tasks requiring deep, specialized processing of images and text. 

However, this method can be computationally complex in terms of having more than one set of encoders and an additional step for the integration (often requiring some kind of sophisticated attention mechanism to align the modalities effectively).

\begin{figure*}[t]
    \centering
    \includegraphics[width=\textwidth]{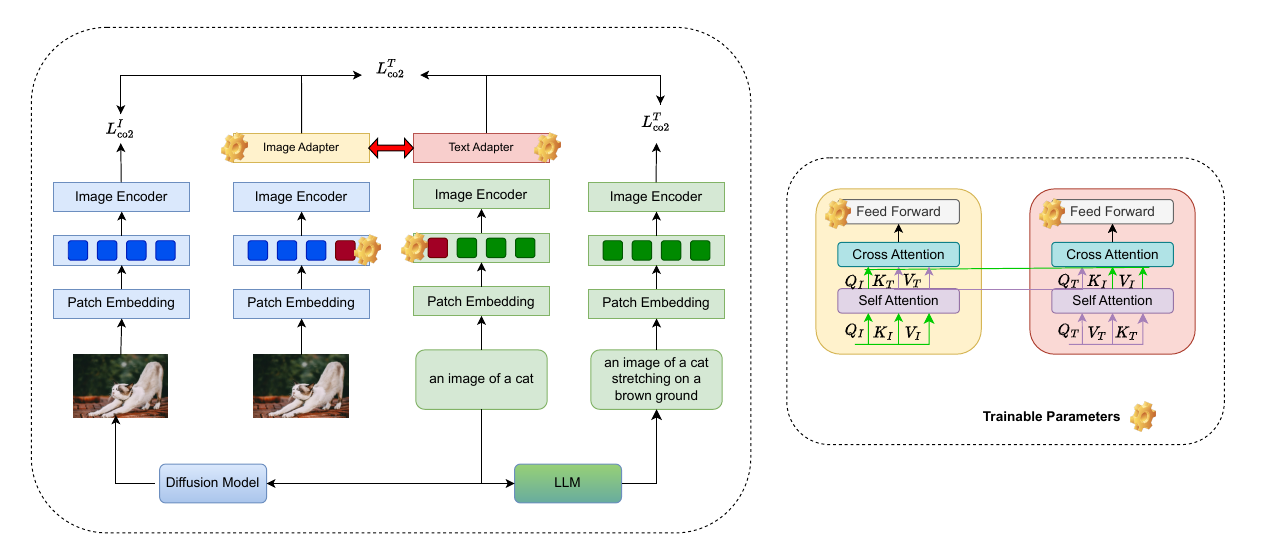}
    \caption{The APoLLo framework provides a unified approach to multi-modal adapter and prompt learning for Vision-Language Pretraining (VLP) models. It incorporates both image (yellow) and text (red) adapters, which are connected via cross-modal attention mechanisms to enhance alignment between the two modalities. Each modality processes augmented inputs: text generated by large language models (LLM) and images synthesized by text-conditioned diffusion models. This cross-modal interaction improves the coherence and performance of multi-modal tasks \cite{chowdhury2023apollo}.}

    \label{fig:apollo}
\end{figure*}

\subsection{\textcolor{black}{Existing Lightweight VLM Models}}

Vision-language models have advanced significantly in recent years, offering capabilities that span across various domains such as image classification, autonomous driving, UI understanding, and more. Table \ref{tab:comparison} shows a comparision between some of the lightweight VLMs, where here we discuss some of the available VLMs, especially those lightweight:

\textbf{ViTamin} is a vision-language model for scalable applications emphasizing image classification and open-vocabulary detection. It uses a Vision Transformer (ViT) base and the CLIP framework, achieving improved zero-shot performance on ImageNet while remaining small. ViTamin processes large datasets, making it suitable for visual recognition tasks and automatic visual description \cite{Chen_2024_CVPR}.

\textbf{LINGO-2}, developed by Wayve, extends vision-language-action models for autonomous driving. It combines visual input, natural language, and action sequences to generate driving behaviors and textual commentaries, increasing explainability. Using a multimodal encoder-decoder architecture, the lightweight 5-billion-parameter model achieves real-world and simulation-capable performance \cite{wayve2024lingo}.

\textbf{InstructBLIP} advances vision-language modeling through instruction tuning, transforming datasets into instruction-following formats. Built on BLIP-2, it surpasses prior state-of-the-art in tasks like question-answering and image captioning, using a Query Transformer for improved adaptability and performance \cite{instructblip2024}.

\textbf{RAVEN} integrates a base VLM with retrieval-augmented frameworks for general-purpose vision-language tasks, excelling in VQA and captioning. Its CLIP-based encoder and transformer decoder enable fine-tuning without retrieval-specific parameters, supporting diverse multimodal applications \cite{raven2024}.

\textbf{ScreenAI} focuses on understanding UIs and infographics through a multimodal encoder-decoder framework. Extending PaLI and incorporating pix2struct's patching strategy, it excels in UI navigation, question-answering, and summarization, leveraging annotated screenshots and infographics \cite{screenai2024}.

\textbf{ALLaVA} uses synthetic data from GPT-4V, employing a captioning-then-QA pipeline with a pre-trained vision encoder and small language model. Fine-tuning on synthesized datasets improves comprehension and reduces hallucinations, achieving strong performance with fewer parameters \cite{allava2024}.

\textbf{Xmodel-VLM}, a lightweight vision-language model for consumer devices, pairs a CLIP ViT-L/14 visual encoder with Xmodel-LM 1.1B, achieving low computational cost and competitive performance on benchmarks \cite{xmodelvlm2024}.

\begin{figure*}[t]
    \centering
    \includegraphics[width=0.8\textwidth]{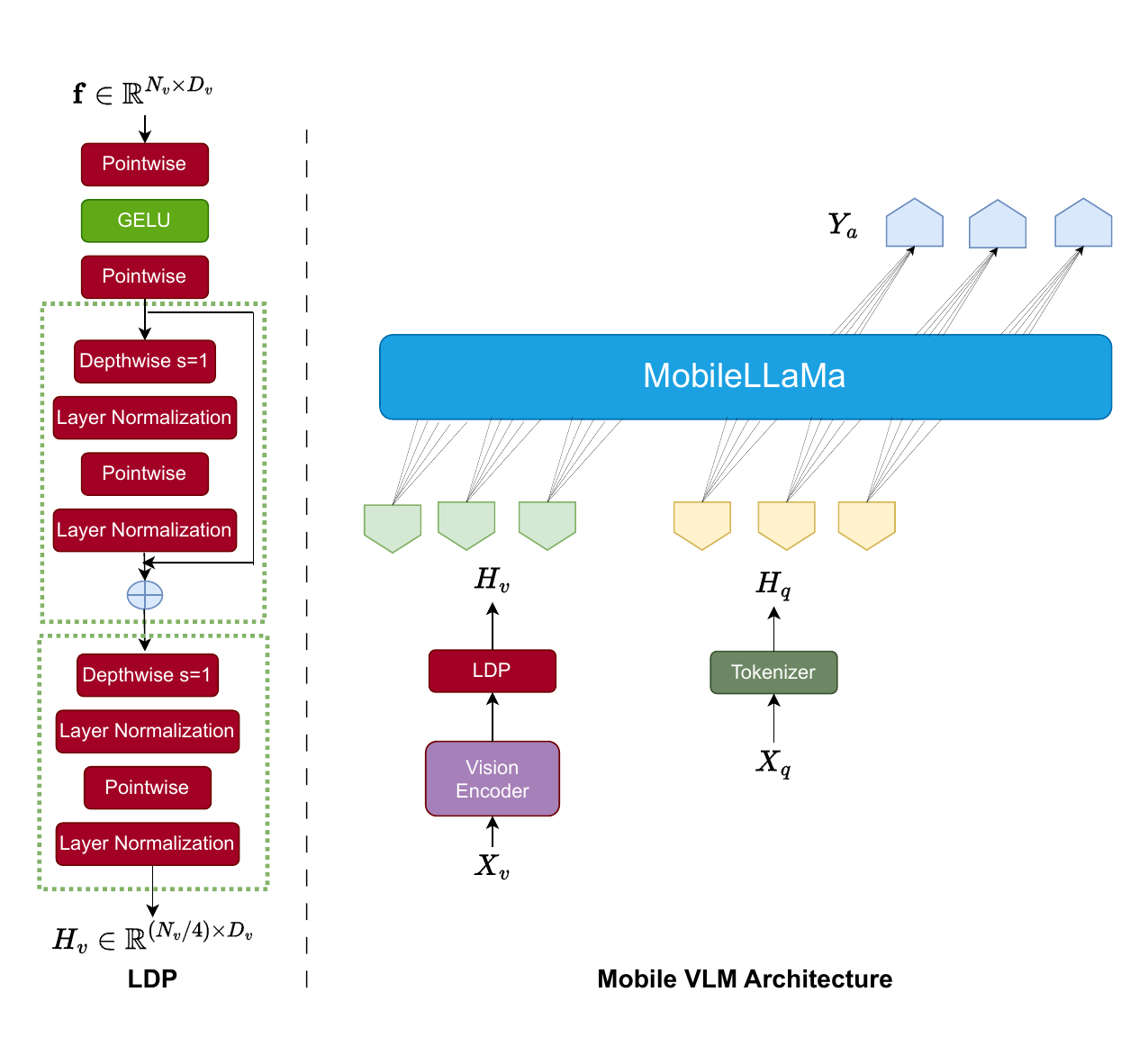}
    \caption{The MobileVLM architecture. Inputs include visual data $X_v \in \mathbb{R}^{N_v \times D_v}$ and textual queries $X_q$, where $N_v$ is the number of visual tokens and $D_v$ is the visual feature dimension. These are processed by a vision encoder and tokenizer, respectively, producing hidden states $H_v \in \mathbb{R}^{(N_v/4) \times D_v}$ and $H_q$. The visual features are passed through a Lightweight Downsample Projector (LDP), which efficiently compresses the input using depthwise and pointwise convolutions. The resulting features $H_v$ and $H_q$ are then fed into MobileLLaMA, a compact vision-language model, which generates the final response $Y_a$.\cite{mobilevlmv2}.}
    \label{fig:mobilevlm}
\end{figure*}

\textbf{MobileVLM V2}, optimized for mobile devices, incorporates a Lightweight Downsample Projector (LDPv2) to reduce visual tokens and speed up inference. Its MobileLLaMA architecture excels in fast, reliable multimodal processing \cite{mobilevlmv2}. Figure \ref{fig:mobilevlm} illustrates the basic model architecture of mobileVLM.

\textbf{LightVLP} adopts the Gated Interactive Masked AutoEncoder architecture for lightweight pre-training. Its multimodal encoder aligns visual and textual inputs efficiently, enabling high-quality outputs with fewer parameters \cite{lightvlp2024}.

\textbf{EM-VLM4AD}, designed for VQA in autonomous driving, combines multi-view image embedding with a gated pooling attention mechanism and a scaled-down T5 language model. It achieves strong performance in perception and planning tasks \cite{emvlm4ad2024}.

\textcolor{black}{
\textbf{EfficientVLM} is a vision-language model designed for efficiency, utilizing knowledge distillation and modal-adaptive pruning to reduce size and enhance speed. With 93 million parameters, it achieves 98.4\% of its teacher model's performance while accelerating inference by 2.2 times, demonstrating strong results in tasks like visual question answering and image retrieval \cite{efficientvlm2024}.}

\textcolor{black}{
\textbf{Unified-IO} is a versatile model capable of handling a wide array of tasks across vision, language, and multimodal domains. By converting diverse inputs and outputs into a unified sequence of discrete tokens, it employs a single transformer-based architecture trained on over 90 datasets. This approach enables Unified-IO to perform tasks such as pose estimation, object detection, and question answering without task-specific fine-tuning \cite{unifiedio2024}.}

\textcolor{black}{
\textbf{MiniVLM} is a compact vision-language model focused on efficiency. It integrates a Two-stage Efficient feature Extractor (TEE) and a transformer-based fusion module, resulting in a 73\% reduction in model size and a 94\% decrease in inference time compared to larger counterparts. Despite its smaller size, MiniVLM retains 94-97\% of the accuracy on various vision-language tasks, making it suitable for edge applications \cite{minivlm2022}.
}

The lightweight VLMs discussed show efficiency and specialization but face challenges in robustness, adaptability, and generalization. Future work should focus on adaptive learning mechanisms, enhanced transfer learning, and dynamic fusion strategies to improve performance in diverse domains and ensure transparency and interpretability for critical applications like healthcare and autonomous driving.

\subsection{\textcolor{black}{Efficient Fine-Tuning Methods for Vision-Language Models}}

Proper fine-tuning mechanisms are critical when adapting large-scale VLMs to downstream tasks with limited computational budgets. Due to their effectiveness in alleviating resource burden related to retraining and full fine-tuning of large models, these techniques have become increasingly popular. This section describes a few diverse lines of research on efficient fine-tuning that emerged in recent years, centering on the topics of prompt-based methods and adapter-based methods.

\subsubsection{Fine-tuning with Prompts}
Their methodology for practicing a specific task with few parameter updates is to shape the input in such a way as to activate the pre-trained model’s capacity, known as prompt-based fine-tuning methods.

\textbf{a. Prompt Tuning:} Creating prompts to prompt the model to produce task-appropriate outputs. Prompts can be hard (discrete text) or soft (continuous vectors). Hard prompts refer to fixed text templates that you include in your input; soft prompts are the continuous embedding of learned vectors injected into your input sequence. CoOp (Context Optimization) and CoCoOp (Conditional Context Optimization) apply learnable soft prompts to enhance the adaptability of the model across varied image recognition tasks \cite{zhou2022learning, zhou2022conditional}.

\textbf{b. Prefix Tuning:} Prefix tuning introduces continuous task-specific vectors (prefixes) to the input of each transformer layer. These prefixes act as virtual tokens, guiding the model's attention mechanism. Lester et al. demonstrated that prefix tuning could achieve competitive performance with minimal additional parameters by adding prefixes to the transformer layers without modifying the original model weights \cite{lester2021power}.

\textbf{c. P-Tuning:} P-tuning extends prompt tuning by using a trainable prefix of virtual tokens that guide the model to focus on task-relevant information. This method is particularly effective in few-shot learning scenarios, where it significantly improves the model's performance with limited data \cite{liu2021ptuning}.

\textbf{d. Prompt Tuning for Vision-Language Models:} Techniques like DenseCLIP and ProDA have been developed to extend prompt tuning specifically for vision-language tasks. These methods use prompt-based learning to align visual and textual features more effectively, achieving performance comparable to full fine-tuning \cite{rao2022denseclip, wang2021proda}.

\subsubsection{Adapter-Based Fine-Tuning}
Adapter-based methods introduce lightweight, task-specific modules into the pre-trained model, allowing efficient adaptation without full model fine-tuning.

\textbf{a. Adapter Modules:} Adapters are small feed-forward networks inserted between the layers of the pre-trained model. They enable task-specific learning by adjusting only the adapter parameters while keeping the original model weights frozen. Houlsby et al. demonstrated that adapter modules could achieve performance comparable to full fine-tuning with significantly fewer trainable parameters \cite{houlsby2019parameter}.

\textbf{b. LoRA (Low-Rank Adaptation):} LoRA reduces the number of trainable parameters by decomposing the weight updates into low-rank matrices. This method allows efficient adaptation of large models with a minimal computational footprint. Hu et al. showed that LoRA could achieve substantial parameter efficiency while maintaining high performance on various downstream tasks \cite{hu2021lora}.

\textbf{c. Parallel Adapter Networks:} Parallel adapters introduce additional parallel pathways in the transformer architecture, allowing for efficient multi-task learning. Pfeiffer et al. proposed AdapterFusion, which combines multiple adapter modules trained on different tasks, enabling the model to leverage shared knowledge across tasks \cite{pfeiffer2021adapterfusion}.

\textbf{d. Task-Specific Adapters:} Techniques like VL-Adapter and Clip-Adapter have been developed to provide efficient task-specific fine-tuning for vision-language tasks. These adapters are designed to handle the unique requirements of multimodal data, improving performance while minimizing computational costs \cite{sung2021vl, gao2021clip}.

\textbf{e. Hybrid Methods:} Some recent approaches combine prompt-based and adapter-based methods to leverage the advantages of both. APoLLo (Adaptive Prompt Learning) integrates prompts and adapters to achieve efficient and robust fine-tuning for vision-language models \cite{chowdhury2023apollo}. Fig.\ref{fig:apollo} explains APoLLo framework for fine-tuning VLMs.

\section{VLMs for Edge Networks}

\begin{figure*}[t]
    \centering
    \includegraphics[width=0.9\textwidth]{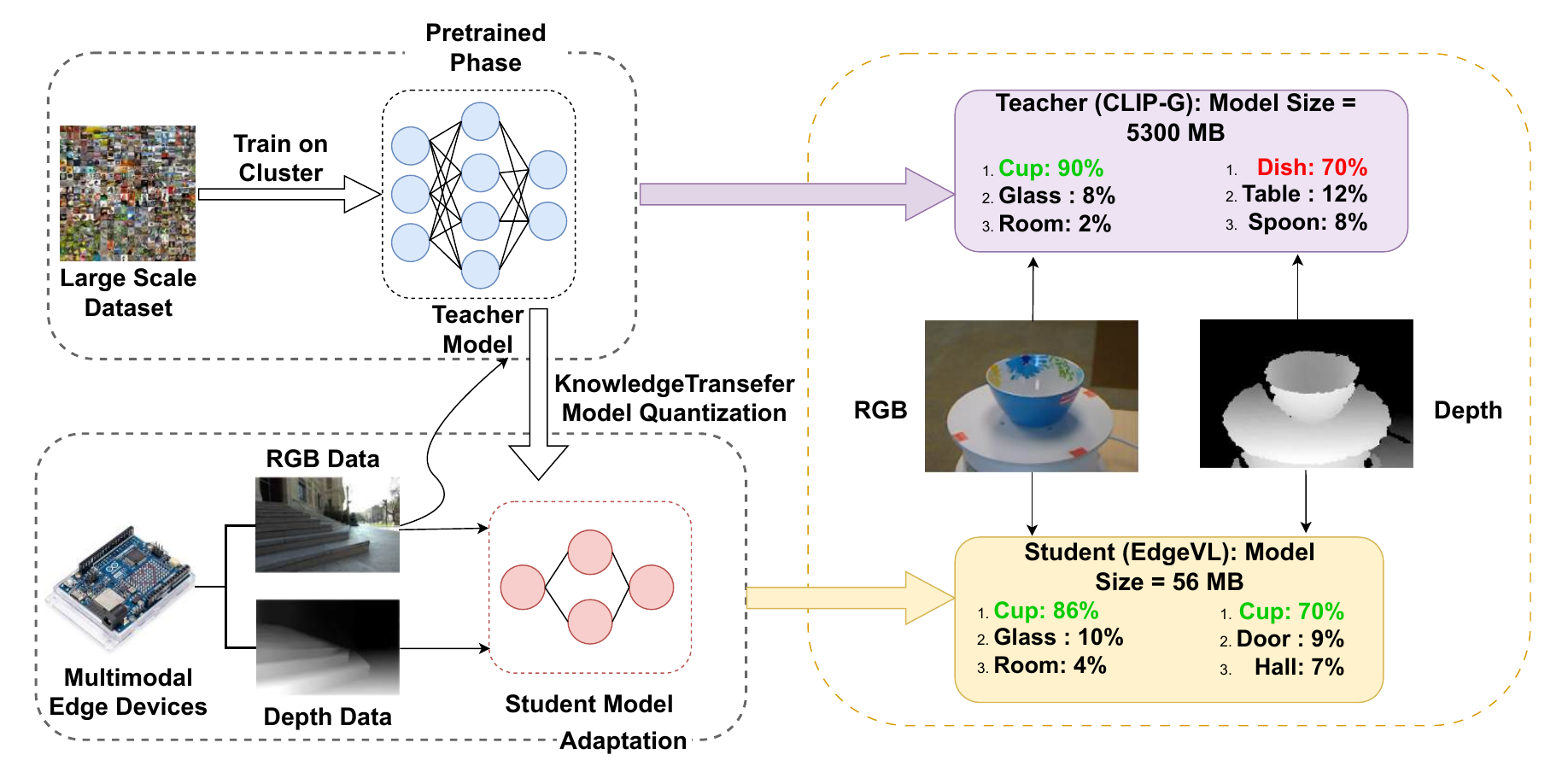}
    \caption{\textcolor{black}{The challenge of adapting large vision-language models to edge devices across different visual modalities. In this example, a resource-constrained cleaning robot equipped with RGB and depth cameras is used. The robot generates RGB-depth image pairs without scene labels. Using the pre-trained image encoder from CLIP as the teacher, the EdgeVL framework transfers knowledge to a smaller student encoder. This process requires no labels or human intervention, enabling the student model to directly process RGB or depth images for open-vocabulary scene classification on the device. EdgeVL distills the knowledge from the pre-trained visual encoder to the student model. In stage 2, it first fake-quantizes the pretrained student model, then uses contrastive learning to refine the student model \cite{edgevl2024}}.
}
    \label{fig:edgevl}
\end{figure*}

Edge devices form an important layer in the IoT architecture and are stationed at the periphery of a network. This allows for real-time insights as they process, store, and compute data locally, transferring less data to potential server farms for processing. The main reasons to deploy edge devices are to deal with lower latency (reduced response time, less significant temporal variability), less usage of data bandwidth, and improved data privacy (sensitive data handling locally) \cite{khan2024resourceICC}.

There are conventional edge devices and intelligent edge devices. Examples of regular edge devices are routers and switches that control the flow of data between the networks with low computation power \cite{alansi2021intelligence}. On the contrary, intelligent edge devices (e.g., IoT gateways and smart cameras) have richer processing abilities to accomplish machine learning inference or data analytics tasks \cite{alizadeh2016authentication}. Mobile devices employ hardware such as System-on-Chip (SoC), Graphics Processing Units (GPUs), and specific processors, making it easier to run complex algorithms on less power \cite{karthikeyan2019applications}.

Below are the key features of edge devices:

\begin{itemize}
\item \textbf{On-Premises Processing}: Local data can be processed at the edge to facilitate rapid data analysis, computation, and feedback without relying on external cloud systems \cite{shiyun2021primer}.
\item \textbf{Autonomy and Low Latency}: These devices provide autonomous decision-making abilities, which are highly necessary for use cases such as self-operated vehicles and manufacturing \cite{shah2020sdn}.
\item \textbf{Higher Security and Privacy}: Data processed at the edge limits exposure of sensitive information, resulting in a higher level of data security and privacy \cite{alansi2021intelligence}.
\item \textbf{Versatile}: Edge devices can be used for a diverse range of applications, including smart cities, industry monitoring, healthcare, and consumer electronic applications \cite{xiao2019edge}.
\end{itemize}

Edge devices have specific technical characteristics depending on the application. SoCs are very much used in the IoT gateways for effective data processing between balanced computational time and energy efficiency. On the other hand, for heavy computing tasks, such as real-time image processing in smart cameras, GPUs or special processors like Application-Specific Integrated Circuits (ASICs) may be used \cite{karthikeyan2019applications}.

\subsection{Existing Low Complexity VLMs}
IoT is a network of devices that connect to the internet to collect, transmit, and analyze data. These may include sensors, smart appliances, wearables, and industrial devices. With the combination of IoT systems with advanced technologies such as Large Language Models (LLMs) and VLMs, sophisticated applications have been achieved, enabling automation, decision-making, and user interaction. In contrast, IoT systems generally consist of three essential layers: perception layer, network layer, and application layer \cite{shiyun2021primer, shah2020sdn}. Sensors and actuators responsible for data collection and control actions are also part of the perception layer.

\textbf{Data Network Layer:} Securing communications between devices and centralized systems, the network layer makes sure that the data can be sent from one device to another or to a centralized system and is often based on protocols such as Wi-Fi, Bluetooth, or LPWAN. Here, the application layer operates over wired or wireless mediums to process and analyze the data to provide useful insights and services to end users \cite{shiyun2021primer, shah2020sdn}.

VLMs are models that combine visual understanding with human-like text generation or understanding to enhance the human-machine IoT interaction experience. The integration of LLMs and VLMs has also resulted in Generative IoT (GIoT) systems, which support the automation of complex tasks, enrich user interactions, and enable real-time decisions \cite{efficient2024prompting, llmind2023}.

\textcolor{black}{
While all edge-deployed Vision-Language Models (VLMs) must be lightweight, not all lightweight VLMs are specifically optimized for edge deployment. Lightweight VLMs focus on reducing model size, computational cost, and latency while maintaining strong performance, but they may still rely on cloud-based processing. In contrast, edge-specific VLMs are designed with additional constraints such as energy efficiency, low-latency operation, and privacy-preserving inference, making them more suitable for real-time applications on resource-constrained devices.}

\textcolor{black}{
Several models have been developed for edge environments, with varying degrees of optimization. One such example is \textbf{EdgeVL}, a framework designed to adapt large-scale models for edge deployment. Unlike general lightweight VLMs, which primarily aim to reduce size and complexity, EdgeVL incorporates dual-modality knowledge distillation and quantization-aware contrastive learning to align representations between a large teacher model, such as CLIP, and a compact student model. This enables efficient processing of RGB and non-RGB images without manual annotation, improving adaptability across vision tasks. The framework achieves significant improvements in accuracy and a substantial reduction in model size, making it highly suitable for real-world edge applications~\cite{cai2024selfadapting}.}

\textcolor{black}{
Other models, such as MiniVLM and LightVLP, represent general-purpose lightweight VLMs that, while optimized for efficiency, are not explicitly designed for edge deployment. MiniVLM reduces visual feature extraction time by 95\% compared to baseline models and achieves competitive performance on various downstream tasks~\cite{wang2020minivlm}. LightVLP employs a gated interaction mechanism to handle noise in aligned image-text pairs, resulting in efficient and effective vision-language pre-training~\cite{xu2023lightvlp}. However, these models do not incorporate edge-specific techniques such as on-device inference, federated learning, or adaptive computation, making them more suited for cloud-assisted environments or mobile applications rather than fully autonomous edge computing.}

\textcolor{black}{
Selecting the appropriate model depends on deployment constraints. General lightweight models may be effective in low-power cloud-assisted settings, while edge-specific models integrate optimizations that enable local inference with minimal latency and energy consumption, making them essential for IoT, autonomous systems, and privacy-sensitive applications.}

\par

\textbf{Moondream2}: Moondream2 is an open-source, lightweight vision-language model (VLM) optimized for mobile and edge devices. With 1.8 billion parameters, it requires under 5GB of memory, making it deployable on low-cost, single-board computers such as Raspberry Pi. Its architecture is designed for efficiency, enabling real-time image recognition and understanding capabilities. This model is suitable for applications such as security and behavioral analysis, showcasing its utility in low-resource environments \cite{moondream2024}.

\textbf{VILA (Visual Language) model}: VILA focuses on pre-training techniques optimized for efficient edge deployment. The model employs interleaving data and instruction fine-tuning to maintain high performance while reducing computational demands. It is adaptable to various hardware, including devices like Jetson Orin. VILA also emphasizes multi-modal pre-training, enhancing in-context learning and multi-image reasoning capabilities \cite{vila2024}.

\textbf{MobileVLM V2}: Building upon the MobileLLaMA series, MobileVLM V2 emphasizes lightweight design for edge deployment. It introduces a novel Lightweight Downsample Projector (LDPv2) that improves vision-language feature alignment with minimal parameters. This approach involves pointwise and depthwise convolutions, along with a pooling layer to compress image tokens. MobileVLM V2 achieves significant reductions in model size and computational requirements, making it ideal for real-time applications on resource-constrained devices \cite{mobilevlmv2}.

\textbf{EDGE-LLM}: EDGE-LLM is a framework designed to adapt large language models for efficient deployment on edge devices. It addresses computational and memory overhead challenges through techniques like efficient tuning and memory management. This model supports continuous and privacy-preserving adaptation and inference, offering a robust solution for deploying VLMs in sensitive and resource-limited environments \cite{edgeLLM2024}.

\textbf{Vision Transformer Models}: Recent advancements in vision transformers have been adapted for mobile and edge devices to maintain high accuracy with minimized model size. Techniques such as token pruning, quantization, and the introduction of convolutions in transformers (e.g., CvT and TinyViT) have been explored. These models cater to tasks like object detection and instance segmentation, highlighting the versatility of vision transformers in edge applications \cite{transformer2024}.

\subsection{Deployment of VLMs on Edge Devices}

\begin{figure}[t]
    \centering
    \includegraphics[width=0.45\textwidth]{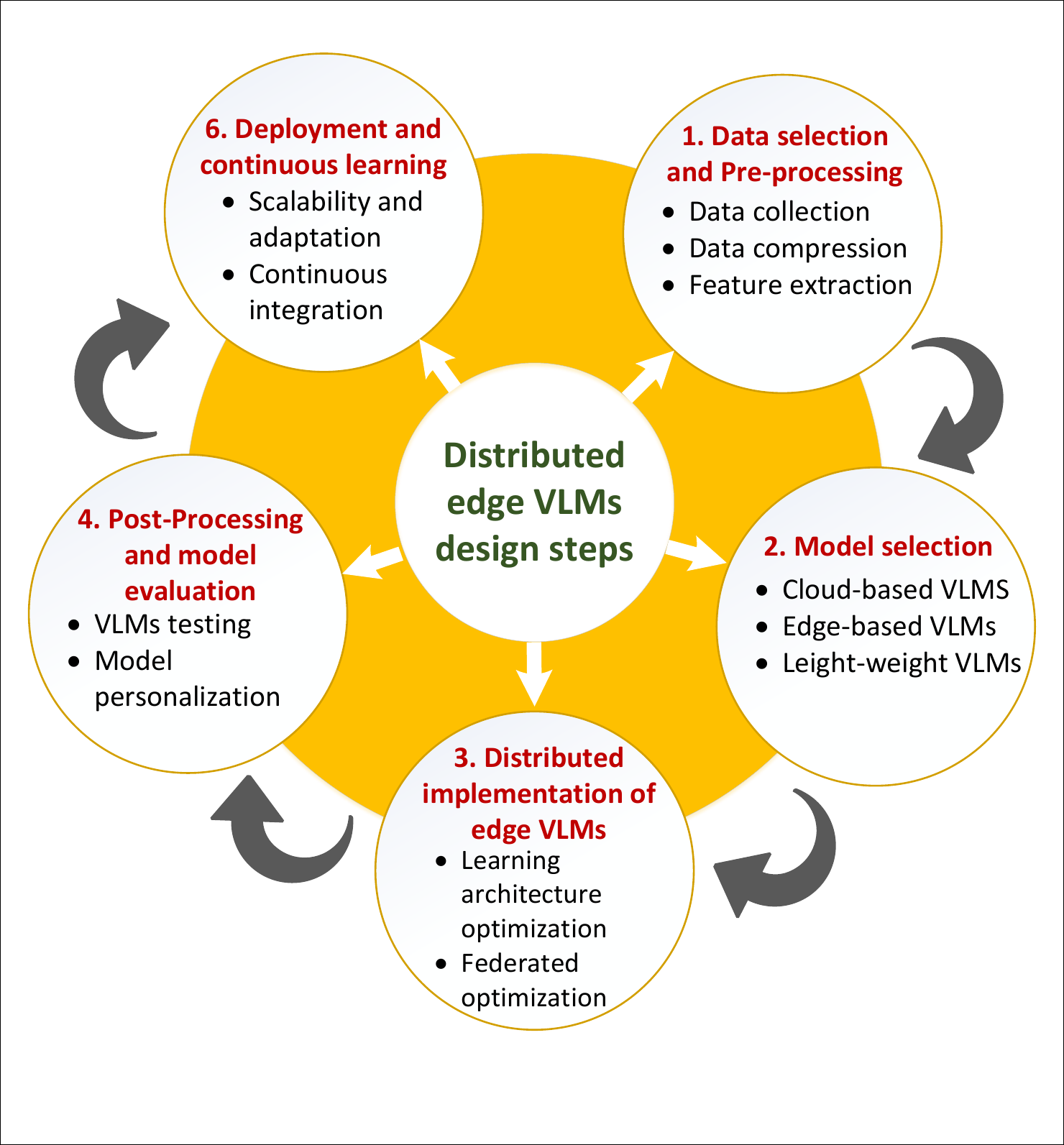}
    \caption{\textcolor{black}{Design Process of Distributed Edge VLMs.}}

    \label{fig:design}
\end{figure}

Deploying VLMs on edge devices requires several key steps for efficiency, as illustrated in Fig.~\ref{fig:design}:

\subsubsection{Data Selection and Pre-processing}

Effective pre-processing is crucial for optimizing performance, especially with heterogeneous data across edge devices. It begins with \textbf{Data Collection}, where various data types (images, text, etc.) are gathered from multiple sources. Data is then segmented into Edge-Appropriate and Cloud-Appropriate categories. Edge-appropriate data is simpler and can be processed in real-time on devices using lightweight models, while cloud-appropriate data requires more complex processing and resources \cite{Shi2021DataSelection,Souza2024BICFL}.

In \textbf{Feature Extraction and Selection}, relevant features are derived from raw data to support efficient processing by edge models. Feature selection helps determine which features are processed locally and which are sent to the cloud, often using heuristics or lightweight models \cite{Sha2024EdgeFL,Hong2024Auction}.

\textbf{Data Compression} further reduces bandwidth for edge-cloud transmission using methods like quantization, dimensionality reduction, and image compression. Local pre-processing helps minimize transmitted data volume \cite{Souza2024BICFL,Sha2024EdgeFL}. Advanced methods like Asynchronous Aggregation and Cluster Pairing introduce intermediate edge servers to aggregate local models asynchronously, reducing communication overhead and accelerating convergence \cite{Sha2024EdgeFL}. Bioinspired Computing (BIC) algorithms, such as PSO and Genetic Algorithms, help tackle communication costs and system heterogeneity by optimizing resource allocation and data partitioning \cite{Souza2024BICFL}. The Synchronous-Asynchronous Hybrid Update Strategy mitigates staleness from Non-IID data by combining local updates with global synchronization, improving accuracy and reducing idle time \cite{Sha2024EdgeFL}. These pre-processing strategies are vital for efficient federated learning in distributed, resource-constrained settings.

\subsubsection{Model Choice on Edge and Cloud}

Performing better when one tries to deploy models on edge or cloud must involve a thoughtful choice as a tradeoff between performance and resources. Deciding where to process information is influenced by the computational capacity of edge devices, task complexity, and the need for real-time processing, among others.

\textbf{Edge-Appropriate Models} are lightweight models meant to work under low computational power constraints. These models should also be computationally less expensive and, therefore, capable of performing inference in real-time on more straightforward tasks. Examples include MobileNet and SqueezeNet, which have fewer parameters and optimized architecture for low-power environments \cite{Howard2017MobileNets,Sandler2018MobileNetV2}. 

\textbf{Cloud-Appropriate Models}, on the other hand, refer to deeper and resource-heavy architectures such as ResNet-50, BERT, as well as other large transformer-based architectures that require a considerable amount of computing resources to maintain but offer improved accuracy and broader analysis capabilities when processed from cloud environments \cite{He2016ResNet,Devlin2019BERT}.

Cloud-native models can also be very complex and resource-intensive. Larger models designed for advanced tasks need high computational power and significant memory resources. These models include ResNet-50, BERT, or even larger transformer architectures, which require more resources than are suitable for edge execution but can achieve better accuracy and more profound analysis when executed in cloud environments \cite{He2016ResNet,Devlin2019BERT}.

\begin{figure}[ht!]
    \centering
    \includegraphics[width=0.45\textwidth]{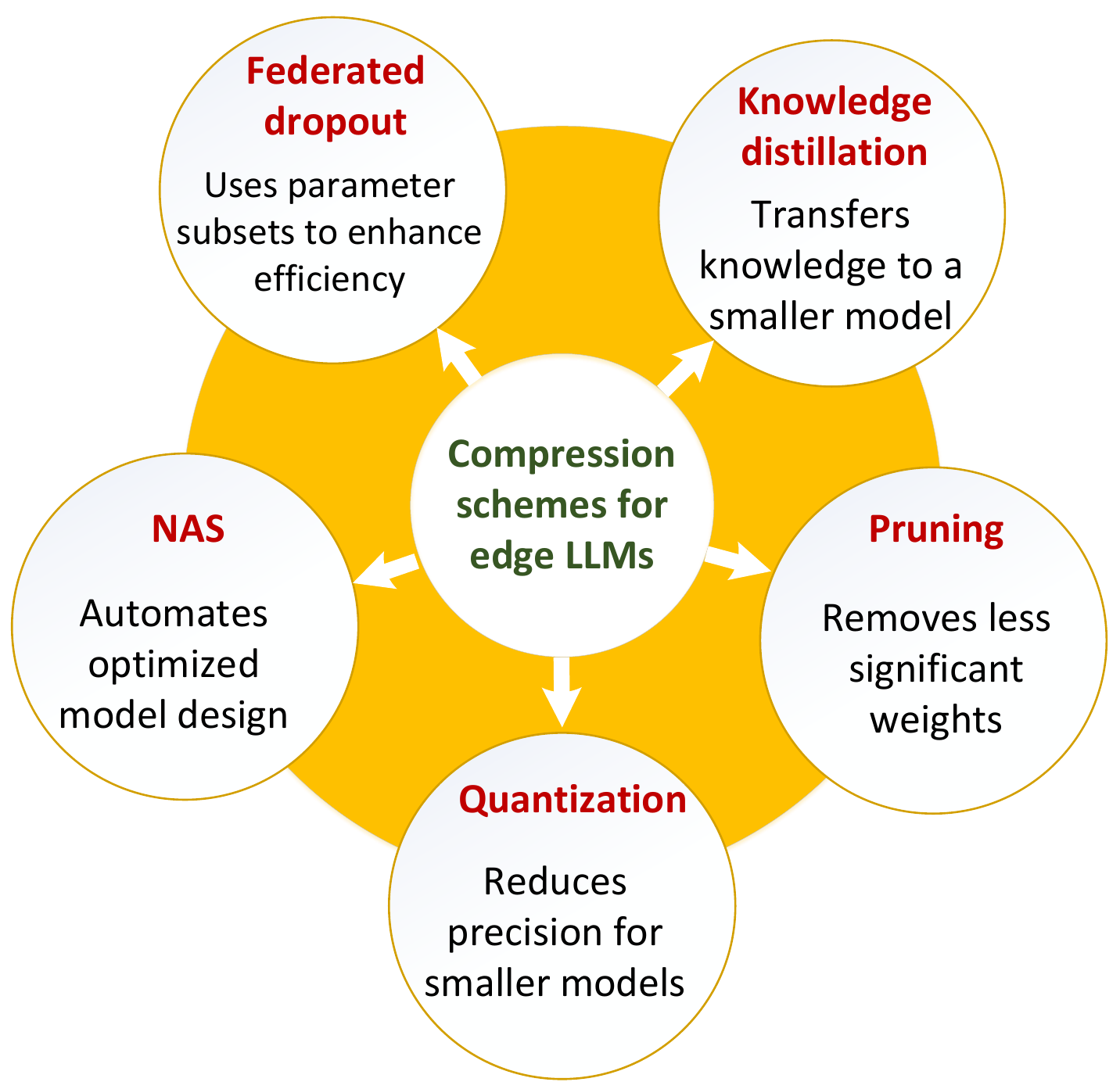}
    \caption{\textcolor{black}{An Overview of Compression Schemes for VLMs.}}
    \label{fig:compression}
\end{figure}

\textbf{Compression Techniques}: Model compression enables deploying complex models on resource-constrained edge devices by reducing size and computational demands without major performance loss. Fig.~\ref{fig:compression} summarizes key techniques, including quantization, pruning, and knowledge distillation. In \textbf{Quantization}, weights and activations are reduced from 32-bit floating-point to lower-bit formats (e.g., 16-bit or 8-bit integers), significantly reducing size and computation \cite{Jacob2018Quantization}. \textbf{Pruning} removes less significant neurons, weights, or layers, cutting model size and inference time \cite{han2016deepcompression}. \textbf{Knowledge Distillation} transfers knowledge from a large teacher model to a smaller student model, achieving comparable performance with fewer parameters, making it ideal for edge deployment \cite{Hinton2015Distillation}.

Other advanced methods further improve compression and efficiency. \textbf{Neural Architecture Search (NAS)} automates the design of efficient architectures by exploring a search space optimized for edge use \cite{Elsken2019NAS}. \textbf{Layer-wise Adaptive Rate Scaling (LARS)} adjusts learning rates across layers during training, and can be combined with other methods to fine-tune performance \cite{You2017LARS}. \textbf{Federated Dropout} uses different subsets of a model’s parameters during training in a federated setup, reducing communication costs and yielding compact, efficient models for edge deployment \cite{Caldas2018FederatedDropout}. These methods support effective deployment across edge and cloud environments, balancing performance with operational constraints.

\textcolor{black}{
\textbf{Balancing Model Compression and Accuracy}}

\textcolor{black}{
Implementing model compression techniques such as pruning, quantization, and knowledge distillation involves a trade-off between computational efficiency and accuracy. Aggressive compression reduces model size and inference time but can degrade performance. To balance this, the following strategies are commonly employed:}

\begin{itemize}
    \item \textit{\textcolor{black}{Incremental Compression:}}
    \textcolor{black}{
    Gradually applying compression enables accuracy monitoring at each step, helping find an optimal trade-off. \cite{chen2024comprehensive} showed that structured pruning applied incrementally achieved substantial size reduction with minimal accuracy loss.}

    \item \textit{\textcolor{black}{Layer-wise Sensitivity Analysis:}}
    \textcolor{black}{
    Assessing layer sensitivity guides which layers can handle more aggressive compression. Less sensitive layers can be pruned or quantized more without major performance loss. \cite{xia2022structured} demonstrated that this approach preserves performance while lowering complexity.}

    \item \textit{\textcolor{black}{Hybrid Compression Techniques:}}
    \textcolor{black}{
    Combining methods like pruning, quantization, and distillation leverages their strengths. Ref.~ \cite{movva2022combining} showed that such combinations reduce size effectively without compromising accuracy.}
    
    \item \textit{\textcolor{black}{Adaptive and Dynamic Compression:}}
    \textcolor{black}{
    Adaptive methods like dynamic quantization adjust precision based on data complexity. Ref.~\cite{li2022dq} proposed a joint distillation and quantization method that dynamically adjusts parameters for efficient compression.}

    \item \textit{\textcolor{black}{Domain-specific Performance Metrics:}}
    \textcolor{black}{
    Defining accuracy thresholds based on application needs ensures compression does not exceed acceptable degradation. Ref.~\cite{chen2024comprehensive} highlighted aligning strategies with domain-specific metrics for practical effectiveness.}
\end{itemize}

\textcolor{black}{
Implementing these strategies requires careful consideration of the specific application and deployment environment to maintain a balance between efficiency and accuracy.}

\subsubsection{VLMs Implementation Across Distribution Locations}

Federated Learning (FL) is an innovative machine learning method that enables the training of models across decentralized devices or servers without requiring raw data exchange. This paradigm resolves critical issues about data privacy, security, and regulatory compliance, as it allows individual data sources (i.e., mobile devices or edge servers) to work together to train a shared global model without the need to share their data, keeping it stored locally. The FL method features multiple rounds of aggregating model updates from all devices involved in the training process, and the new global model is synthesized based on the received updates. It minimizes the possibility of private data breaches and reduces communication costs associated with centralized data processing. Federated Learning demonstrates the most promise in scenarios involving distributed, privacy-sensitive, and heterogeneous data, contributing to its applications in healthcare, finance, and IoT environments \cite{Kairouz2019AdvancesFL,Yang2019FederatedLearning,McMahan2017CommunicationEfficientFL}. State-of-the-art scalable and robust FL systems, coupled with advanced techniques such as model compression (to lower the model size), federated averaging (for communication efficiency), and differential privacy (to enhance efficiency in resource-constrained environments), have further improved FL systems \cite{McMahan2017CommunicationEfficientFL,Caldas2018FederatedDropout}. FL is still maturing, and it will be one of the building blocks of secure and efficient machine learning in the era of ubiquitous computing. 

\par

\paragraph{Model Partitioning}
The setup of FL begins with \textbf{model partitioning}, which defines the architecture of the parent model on the cloud (computation-heavy) and the child model on the edge (lightweight). The Parent-Child Model Setup is key: the child model runs locally with minimal resources, handling simple tasks; the parent model is more complex and handles updates. A hierarchical architecture enables the child model to perform online inference independently with minimal on-device memory, while the cloud model manages updates and intensive computation \cite{Kairouz2019AdvancesFL,Yang2019FederatedLearning}.

The next stage involves delegating tasks. Edge models handle real-time predictions and local processing with low or no latency, while the cloud model manages resource-intensive tasks such as model aggregation, complex analytics, and batch updates. This division allows efficient use of both edge and cloud resources \cite{McMahan2017CommunicationEfficientFL,Sattler2019RobustFL}.

\paragraph{Data Distribution Policy}

Once models are separated, a data distribution policy is applied. Local data handling on the edge supports generating updates or predictions with minimal cloud interaction.

\begin{enumerate}

  \item \textbf{Selective Uploading}: Determines which processed data are essential for cloud upload, sending only significant updates or data requiring deeper analysis \cite{Smith2017FederatedMultiTask,Yurochkin2019BayesianNonparametricFL}. This reduces network usage and enhances system efficiency.

  \item \textbf{Model Update Mechanism}: Maintains synchronization between the edge model and cloud model updates. Local training allows the child model to learn from on-device data without exceeding limited computation capacity.

  \item \textbf{Federated Averaging Algorithm}: Updates from edge devices are averaged on the cloud to refine the global model, enabling learning from non-independent data sources while minimizing data leakage risk \cite{McMahan2017CommunicationEfficientFL,Caldas2018FederatedDropout}.

  \item \textbf{Model Synchronization}: Ensures the child model incorporates global updates via periodic syncing. The cloud returns updated parameters to edge devices, enabling local models to benefit from collective learning and maintain accuracy \cite{Sattler2019RobustFL,Konečný2016FederatedOptimization}.

\end{enumerate}

\textbf{\textcolor{black}{Privacy and Security Considerations}}

\textcolor{black}{
Deploying VLMs on edge devices necessitates robust privacy and security measures to protect sensitive data. Several advanced techniques have been developed to address these concerns:
}

\textcolor{black}{
\textit{Differential Privacy (DP):} DP introduces controlled noise to the data or model updates, ensuring that individual data points cannot be distinguished, thereby preserving privacy. In federated learning scenarios, DP has been effectively applied to protect client data during model training \cite{truex2020ldp}.
}

\textcolor{black}{
\textit{Secure Aggregation:} This technique ensures that the server can aggregate model updates from clients without accessing individual contributions. Methods utilizing multiparty homomorphic encryption have been proposed to achieve efficient and secure aggregation in federated learning \cite{bonawitz2017practical}.
}

\textcolor{black}{
\textit{Homomorphic Encryption (HE):} HE allows computations to be performed directly on encrypted data without decryption, maintaining data confidentiality throughout the process. Implementations like OpenFHE have been developed to support such privacy-preserving computations \cite{openfhe2022}.
}

\textcolor{black}{
Integrating these techniques into the deployment of VLMs on edge devices enhances data privacy and security, addressing potential vulnerabilities in decentralized networks.
}

\subsubsection{Post Processing and Evaluation}

Post-processing and evaluation are critical in the FL system life-cycle to ensure both local and global models perform well across heterogeneous environments. The first step is \textbf{Local Evaluation}, which assesses child model performance on edge devices using local data. Detecting performance degradation indicates when updates from the cloud are needed. Recent approaches use continual learning and on-the-fly performance tracking to adapt models dynamically to changing edge contexts \cite{Yao2022EdgeFL}. As noted by \cite{Wang2021EdgeContinualLearning}, lightweight performance estimation and anomaly detection run locally to monitor behavior and trigger cloud deployment requests.

The feedback loop is integral to local evaluation, helping identify difficult scenarios or data for the edge model. This feedback improves the global model by highlighting areas needing more training. Reinforcement learning and meta-learning techniques are often used to help edge models self-improve over time, also benefiting the global model \cite{Li2021FederatedMetaLearning}.

\textbf{Global Model Evaluation} aggregates updates and data from multiple edge devices in the cloud, allowing a comprehensive assessment across varied environments and user contexts. Federated evaluation frameworks use privacy-preserving techniques to gather performance metrics without exposing user data \cite{Xu2022FederatedEval}. Secure aggregation and differential privacy are key to enabling accurate and private evaluation.

\textbf{Model Tuning on the Aggregation}: The global model is fine-tuned on the server using updates from edge devices, capturing local data nuances and improving generalizability. Advanced strategies like federated hyperparameter tuning and federated Bayesian optimization ensure the global model remains effective across devices and conditions \cite{Zhang2021FederatedHyperparameter}.

\subsubsection{Deployment and Continuous Learning}

Deployment and continuous learning ensure FL models are updated, scaled, and adapted to evolving environments. Recent advances have introduced SOTA methods to enhance robustness and scalability.

\textbf{Model Update Deployment} — Model parameters are frequently pushed from the cloud to edge devices to support ongoing learning. New approaches focus on lightweight updates and differential synchronization to reduce data exchange, minimizing latency and bandwidth use \cite{Liu2024ContinuousFL}. Sparse update techniques like federated dropout send only essential parameters, improving efficiency \cite{Zhang2024SparseUpdates}.

Edge Device Management automates update deployment across devices using orchestration frameworks. Decentralized strategies (e.g., peer-to-peer) reduce server load and improve scalability \cite{Wang2024EdgeDeployment}. Adaptive techniques tailor updates to each device’s capabilities \cite{Chen2024AdaptiveFL}.

A \textbf{Scalable Architecture} is key for growing edge networks. SOTA solutions scale horizontally by adding edge nodes and vertically via cloud resource expansion. Cloud-native tools like Kubernetes and containerization enable dynamic scaling across large device clusters \cite{Xu2023KubeFL}.

Adaptation to New Data is vital for maintaining accuracy as data evolves. Continual learning and federated meta-learning help models adapt without forgetting past knowledge \cite{Yao2024ContinualFL}. Reinforcement learning allows dynamic strategy adjustments in non-stationary environments \cite{Zhu2024RLFL}.

Current Vision-Language Models are difficult to deploy on edge devices due to high resource requirements. Designed for cloud use, they face challenges like latency, poor real-time performance, and power inefficiency. They also struggle with diverse visual modalities and rely on centralized data aggregation, raising privacy and dependency concerns.

To address this, VLMs for edge should use lightweight architectures that balance performance and efficiency. Techniques like quantization, pruning, and knowledge distillation reduce size and computation without losing accuracy. Neural Architecture Search and Federated Learning improve adaptability and privacy in distributed settings \cite{khan2023joint}. Future models must support real-time, multimodal processing and continual learning for dynamic edge environments.

\begin{figure*}[t]
    \centering
   \includegraphics[width=0.8\textwidth]{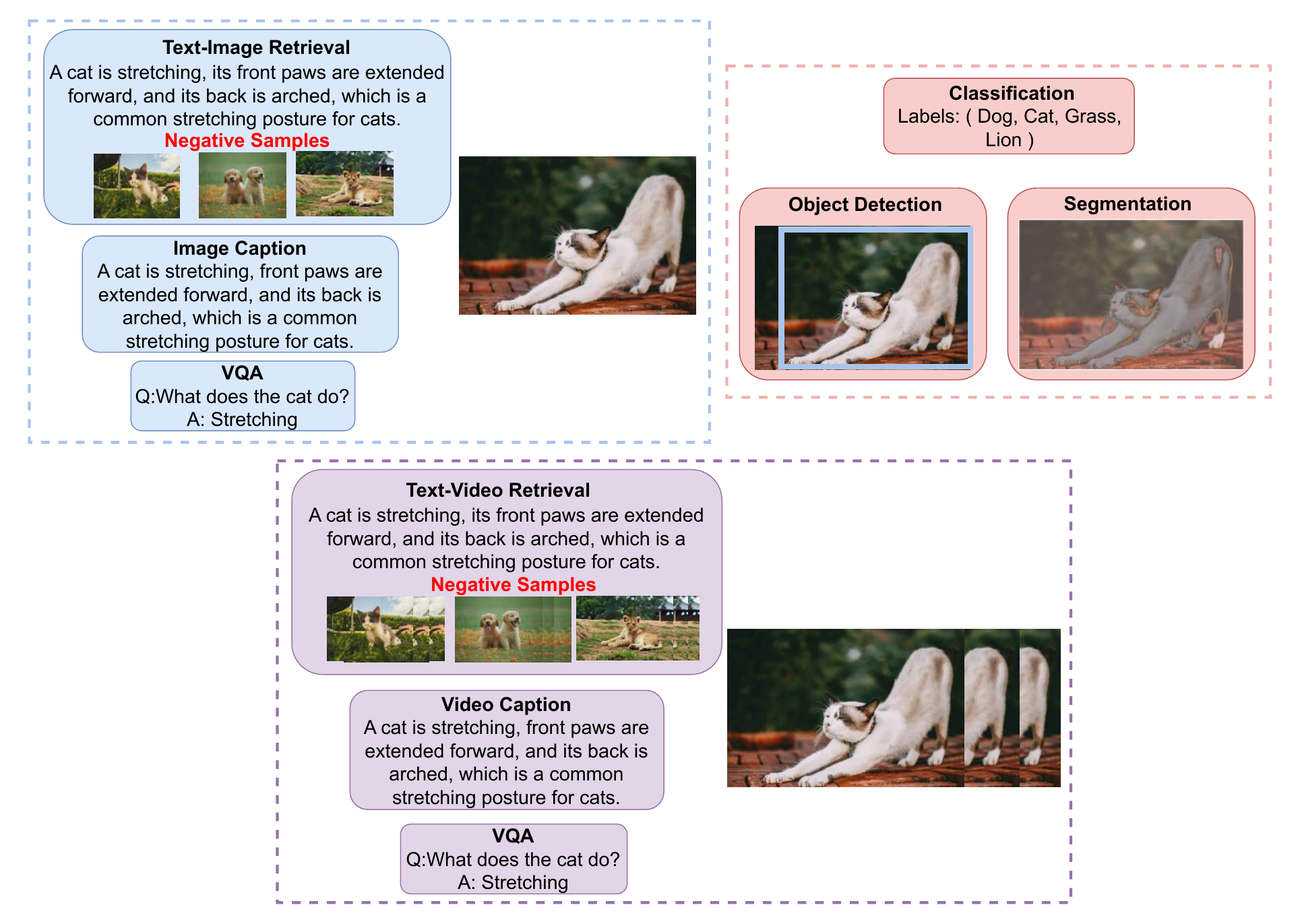}
    \caption{\textcolor{black}{Illustration of representative tasks from three categories of Vision-Language (VL) problems, image-text tasks, vision tasks as VL, and video-text tasks \cite{visionlanguagepretrainingbasicsrecent}.}}
    \label{fig:VL Overview}
\end{figure*}

\section{Recent Advances}

Applications of lightweight VLMs are expanding rapidly across various industries, driven by the need for efficient, on-device multi-modal processing. There are many tasks where VLMs can help, including Text-Image Retrieval, image captioning, Question and answer Classification, object detection, and segmentation, as shown in Figure \ref{fig:VL Overview}. These models are being deployed in autonomous systems, healthcare, surveillance, environmental monitoring, and many other applications. Table \ref{tab:vlm_applications} summarizes the applications we covered in this survey.

\subsection{VLM in healthcare}

VLMs are increasingly important in various medical applications. Their simultaneous processing of visual and textual data allows for more accurate diagnoses and effective medical workflows. Below are some applications of VLMs in the medical domain, particularly focusing on lightweight models designed to work on edge devices.

\textbf{Bilingual Medical Mixture LLM}  
\cite{li2024bimediX} propose BiMediX, the first bilingual medical mixture of experts LLM designed for seamless interaction in both English and Arabic. BiMediX supports various medical tasks, including multi-turn dialogues, multiple-choice questions, and open-ended queries. We developed a semi-automated English-to-Arabic translation pipeline and a comprehensive evaluation benchmark for Arabic medical LLMs. We also present BiMed1.3M, a bilingual dataset of 1.3 million medical interactions, which powers the model's instruction tuning. BiMediX outperforms state-of-the-art models Med42 and Meditron, offering 8-times faster inference, and exceeds Jais-30B on both Arabic and bilingual medical benchmarks.

\textbf{Medical Visual Question Answering (VQA)}  
Integrating VLMs in medical visual question answering tasks can help doctors make faster and more informed decisions. \cite{delbrouck2022vilmedic} developed ViLMedic, a multimodal framework that supports a variety of medical tasks such as visual question answering and radiology report generation. This framework includes multiple pretrained models designed for efficient deployment on edge devices, enabling real-time interaction with medical data.

\textbf{Computer-Aided Diagnosis (CAD)}  
Another key application is computer-aided diagnosis (CAD). \cite{chen2023medblip} proposed MedBLIP, a lightweight VLM designed to analyze 3D medical images and electronic health records for Alzheimer’s diagnosis. The model uses pre-trained image encoders and large language models to provide accurate zero-shot classification for Alzheimer’s disease. This approach demonstrates the viability of VLMs in providing real-time CAD support on resource-constrained devices.

\textbf{Ultrasound-Guided Diagnosis for COVID-19}  
Portable ultrasound devices are critical in diagnosing diseases like COVID-19. \cite{karar2021covid} proposed COVID-LWNet, a lightweight deep learning model designed to classify lung conditions using ultrasound images. The model showed excellent performance on edge devices, making it suitable for field deployment where real-time diagnostic support is essential.

\textbf{Medical Report Generation}  
VLMs have also been used to generate medical reports based on visual data automatically. \cite{yang2023customizing} introduced a system that integrates VLMs with large language models to generate detailed medical reports. This system has shown great potential in automating radiology report generation and improving efficiency and accuracy in clinical settings.

\subsection{VLMs in Environmental Monitoring and Aerial Imaging}

VLMs are increasingly being employed in environmental monitoring, including tasks such as aerial imaging, environmental change detection, and disaster assessment. These models are particularly valuable when deployed on edge devices for real-time monitoring in remote or resource-constrained areas. Below are some of the key applications of VLMs in this domain, with a focus on lightweight models designed for edge deployment.

\textbf{Aerial Imaging for Environmental Change Detection}  
One of the critical applications of VLMs in temporal change detection involves monitoring geographic landscapes to support environmental analysis and urban development planning. To address limitations in capturing dynamic shifts, this study introduces an annotated dataset of video frame pairs to trace evolving geographical features over time. Building on techniques like Low-Rank Adaptation (LoRA), quantized LoRA (QLoRA), and model pruning, models such as Video-LLaVA and LLaVA-NeXT-Video are fine-tuned to achieve high accuracy in tracking and describing land-use transformations. GeoLLaVA \cite{geollava} models demonstrate notable performance gains, achieving a BERT score of 0.864 and a ROUGE-1 score of 0.576, underscoring their enhanced capabilities for precise, temporal environmental monitoring.

\textbf{Vision-Language Models for Climate and Land-Use Analysis}  
\cite{roberts2023satin} introduced SATIN, a multi-task metadataset designed for classifying satellite and aerial imagery using VLMs. The model is optimized for environmental applications like land-use planning and deforestation monitoring. It shows high transfer performance in zero-shot classification tasks, making it effective for rapid deployment in environmental surveys on resource-constrained devices.

\textbf{Vision-Language Navigation for UAVs in Environmental Surveys}  
VLMs are also used in UAV-based environmental navigation. \cite{liu2023aerialvln} introduced AerialVLN, a vision-language navigation model designed for UAVs. This model enables UAVs to navigate complex environments while performing environmental surveys or wildlife tracking. AerialVLN's lightweight architecture makes it ideal for deployment on UAVs, enabling real-time decision-making during aerial environmental surveys.

\textbf{Remote Sensing Change Detection (RSCD)}  
ChangeCLIP is a novel framework for remote sensing change detection (RSCD) that leverages the vision-language model CLIP, which aims to improve the detection of surface changes from bitemporal images \cite{DONG202453}. While traditional methods primarily focus on visual representation, ChangeCLIP integrates multimodal data by reconstructing CLIP to extract bitemporal features and proposing a differential features compensation module to capture detailed semantic changes. Additionally, it introduces a vision-language-driven decoder that enhances image semantics by combining image-text encoding with visual features during decoding. ChangeCLIP achieves state-of-the-art results on five RSCD benchmark datasets: LEVIR-CD (85.20\%), LEVIR-CD+ (75.63\%), WHUCD (90.15\%), CDD (95.87\%), and SYSU-CD (71.41\%).

\begin{table*}[ht!]
\centering
\renewcommand{\arraystretch}{1.2} 
\caption{Summary of Vision-Language Model Applications in Different Domains}
\label{tab:vlm_applications}
\begin{tabularx}{\textwidth}{p{3cm} X} 
\hline
\textbf{Application Domain} & \textbf{Application and Description} \\ \hline

\multirow{5}{3cm}{Medical Domain} 
& Bilingual Medical Mixture LLM: BiMediX enables multilingual medical interactions, improving task efficiency in both English and Arabic \cite{li2024bimediX}. \\  
& Medical Visual Question Answering: ViLMedic supports visual question answering and radiology report generation on edge devices for real-time medical data interaction \cite{delbrouck2022vilmedic}. \\  
& Computer-Aided Diagnosis: MedBLIP provides zero-shot classification for Alzheimer’s disease using 3D images and electronic health records \cite{chen2023medblip}. \\  
& Ultrasound-Guided Diagnosis: COVID-LWNet uses ultrasound images to classify lung conditions, performing efficiently on portable devices \cite{karar2021covid}. \\  
& Medical Report Generation: VLMs are integrated with large language models to automate radiology report generation \cite{yang2023customizing}. \\ \hline

\multirow{4}{3cm}{Environmental Monitoring \\ and Aerial Imaging} 
& Aerial Imaging for Environmental Change Detection: A bi-modal transformer-based VLM helps track and assess environmental changes using aerial imagery \cite{bazi2022bimodal}. \\  
& Climate and Land-Use Analysis: SATIN classifies satellite imagery for tasks like deforestation monitoring, optimized for resource-constrained devices \cite{roberts2023satin}. \\  
& UAV Navigation for Environmental Surveys: AerialVLN enables UAVs to navigate complex environments for wildlife tracking and environmental assessments \cite{liu2023aerialvln}. \\  
& Remote Sensing Change Detection: ChangeCLIP improves surface change detection by integrating multimodal data from bitemporal satellite images \cite{DONG202453}. \\ \hline

\multirow{5}{*}{Autonomous Systems} 
& Autonomous Driving and Transportation Systems: VLMs assist autonomous vehicles by enhancing traffic scene understanding and decision-making \cite{zhou2023vlm_autonomous}. \\  
& Language Prompts in Autonomous Driving: NuPrompt integrates vision-language prompts to improve vehicle response to natural language commands in complex driving scenarios \cite{wu2023languageprompt_autonomous}. \\  
& Real-Time Object Detection in Traffic: A VLM system improves object detection under diverse weather conditions in urban intersections \cite{gao2022citytraffic_autonomous}. \\  
& UAV-Based Autonomous Navigation: AerialVLN supports autonomous navigation for UAVs using visual and language inputs for environmental monitoring \cite{liu2023aerialvln}. \\  
& Disaster Response in Autonomous Systems: Crisscross Vision Transformers process aerial imagery for real-time disaster response decision-making \cite{deng2023crisscross}. \\ \hline

\multirow{4}{*}{Surveillance} 
& Urban Dynamics and Policy Compliance: A vision-language model tracks changes in urban activity and policy compliance using dashcam data \cite{chowdhury2021tracking}. \\  
& Threat Detection : VLMs analyze aerial and ground-level imagery to detect potential threats in public areas \cite{wen2023visionlanguage_remote}. \\  
& Real-Time Activity Monitoring: monitors crowded public spaces, identifying anomalies and suspicious behaviors \cite{du2023success_vqa}. \\  
& Smart City Surveillance: VLMs process multimodal data for real-time even monitoring in urban environments \cite{li2023vision}. \\ \hline

\end{tabularx}
\end{table*}

\subsection{Autonomous Applications of VLMs}

VLMs have significant applications in autonomous systems, particularly in autonomous driving. These models enhance autonomous systems by combining visual and linguistic data for a deeper understanding of the environment. Below are key applications:

\textbf{VLMs in Autonomous Driving and Intelligent Transportation Systems}  
\cite{zhou2023vlm_autonomous} studied VLMs in autonomous driving, showing how integrating visual inputs with natural language processing enhances decision-making for driving safety and efficiency. The paper highlights advances and challenges in object detection, traffic scene understanding, and decision-making.

\textbf{Language Prompts in Autonomous Driving}  
\cite{wu2023languageprompt_autonomous} introduced NuPrompt, a vision-language model using prompts to enhance understanding of natural language commands in driving scenes. With a novel benchmark dataset, NuPrompt demonstrated applications in tracking objects and informed decision-making.

\textbf{Real-Time Object Detection in Urban Traffic}  
\cite{gao2022citytraffic_autonomous} proposed a 2-stage VLM system for traffic object detection and classification under varying weather and traffic conditions. It enhances object recognition in urban intersections, improving safety for vehicles and pedestrians.

\textbf{Autonomous Navigation Using Vision-Language Models}  
\cite{liu2023aerialvln} developed AerialVLN, a VLM for UAV-based navigation. It combines visual and language inputs for tasks like environmental monitoring and search-and-rescue missions, enabling real-time decision-making in complex environments.

\textbf{Crisscross Vision Transformers for Autonomous Disaster Response}  
\cite{deng2023crisscross} introduced a Crisscross Vision Transformer for disaster-prone areas. By processing aerial imagery and textual descriptions, the model supports real-time decision-making in dynamic environments like floods and landslides.

\subsection{Surveillance Applications of VLMs}

VLMs have emerged as transformative tools in surveillance due to their ability to integrate visual and textual information, which enhances real-time monitoring, anomaly detection, and decision-making. These models provide contextual insights into scenes and human behavior, making them essential in various surveillance applications.

In urban surveillance, lightweight VLMs facilitate real-time anomaly detection, facial recognition, and scene analysis, all critical for smart city environments where low latency and immediate response are required~\cite{techhq2024}. In healthcare, VLMs extend to medical imaging diagnostics by analyzing X-rays, MRIs, and related reports to deliver on-device diagnostics, which is vital in remote or field hospital settings with limited cloud connectivity~\cite{cai2024self}. Furthermore, VLMs enhance smart home interactions by processing visual and natural language commands. For example, smart cameras equipped with VLMs can describe scenes, enhancing the user experience with IoT devices~\cite{vik2024moondream}. VLMs also contribute to environmental monitoring and precision agriculture, analyzing satellite and drone imagery to optimize water usage, detect pests, and monitor environmental changes in real-time~\cite{cai2024self}. Retail and e-commerce industries also benefit from VLMs, as they support visual search, recommendations, and inventory management, enabling customers to search products with images on mobile devices~\cite{vik2024moondream}.

\textbf{Tracking Urban Dynamics and Policy Compliance}  
\cite{chowdhury2021tracking} proposed a VLM-based sensing model to monitor urban activity across New York City using dashcam data, generating text-based descriptions to track changes in urban patterns and compliance with social distancing. This approach reduced storage requirements and addressed privacy concerns, making it suitable for large-scale urban surveillance.

\textbf{Surveillance and Threat Detection in Public Spaces}  
\cite{wen2023visionlanguage_remote} examined VLMs in remote sensing for threat detection, focusing on public areas. By combining image captioning and object detection, the model analyzes aerial and ground-level imagery to identify potential threats, such as abandoned objects or suspicious behaviors. The integration of visual data with language descriptions enhances interpretability in security contexts.

\textbf{Monitoring Real-Time Activity in Crowded Areas}  
In 2023, \cite{du2023success_vqa} introduced a VLM for detecting predefined behaviors in crowded environments. This model analyzes visual and textual inputs to flag successful or suspicious activities, making it valuable for crowd management and anomaly detection by providing contextual insights to human operators.

\textbf{Vision-Language Models for Smart City Surveillance}  
Smart city surveillance requires processing multimodal data for efficient decision-making. \cite{li2023vision} developed a model to predict human activity in urban settings by integrating visual and language inputs. Connected to city-wide sensors, this model supports automatic detection of unusual events like unauthorized vehicle entry or loitering, enhancing urban safety.

\section{Open Challenges}
This section presents novel challenges in advancing VLMs for edge applications. Recent surveys have identified several critical issues. Table \ref{tab:vlm_challenges} highlights the challenges we introduce and how they differ from prior discussions. The alignment gap between visual and textual modalities limits performance in tasks like image captioning and question answering~\cite{surveyvisionlanguagepretrainedmodels, visionlanguageintelligencetasksrepresentation}. VLMs also rely on large-scale datasets, which are expensive to develop and restrict access for smaller institutions~\cite{ghosh2024exploringfrontiervisionlanguagemodels}. High computational demands make deployment difficult on resource-constrained devices~\cite{efficientmultimodallargelanguage, SurveyofEfficientFine-TuningMethods}, and VLMs often struggle to generalize across domains, underperforming on novel tasks~\cite{zhang2024visionlanguagemodelsvisiontasks}. Parameter inefficiencies in dual-stream architectures further increase memory usage, making them less suitable for real-time or edge-based applications~\cite{surveyvisionlanguagepretrainedmodels}. These issues call for innovative solutions.

Surveyed papers explore several challenges in deploying LLMs on edge devices. Cai et al.~\cite{cai2024selfadapting} highlight computational demands and dataset reliance, along with the visual-textual alignment gap. Bhardwaj et al.~\cite{surveyintegrationoptimization} focus on optimizing computational load, energy efficiency, and model scalability. Lu et al.~\cite{vliai20survey} discuss generalizing VLMs for edge AI, covering wearables, hardware miniaturization, and usability. Qu et al.~\cite{mobileedgeintelligence} propose mobile edge intelligence to bridge cloud and device-side AI, addressing privacy, latency, and resource constraints. Lin et al.~\cite{pushing6Gedge} examine VLM deployment in 6G edge networks, citing response time, bandwidth, and privacy challenges. Yuan et al.~\cite{generativeinferenceoverview} emphasize energy-efficient inference on mobile devices. Qu et al.~\cite{mobileedgeintelligencemodern} critique on-device LLMs due to limited capacity, advocating edge intelligence to reduce latency and enhance privacy. Chen et al.~\cite{edgegeneralintelligence} explore LLMs in edge intelligence, focusing on adaptive use and throughput challenges for small models. Lee et al.~\cite{visiontransformermobileedge} address adapting Vision Transformers for edge use, emphasizing efficiency and deployment challenges on constrained devices.

\begin{table*}[ht!]
\centering
\renewcommand{\arraystretch}{1.5} 
\caption{Challenges Discussed in Surveys on Vision-Language Models for Edge Devices}
\label{tab:vlm_challenges}
\begin{tabularx}{\textwidth}{p{5cm} X} 
\hline
\textbf{Reference} & \textbf{Challenges Discussed} \\ \hline

Cai et al. (2024) \cite{cai2024selfadapting} 
& High Computational Demands, Large Dataset Dependency, Visual-Text Alignment Issues. \\ \hline

Bhardwaj et al. \cite{surveyintegrationoptimization} 
& Edge Deployment Challenges, High Computation, Energy Efficiency, Scalability. \\ \hline

Lu et al. (2024) \cite{vliai20survey} 
& Generalization, AI in Wearables, Miniaturizing Hardware, User Interface Design. \\ \hline

Qu et al. (2024) \cite{mobileedgeintelligence} 
& Cloud-Edge Gap, Privacy And Latency Issues, Limited Edge Device Resources. \\ \hline

Lin et al. (2023) \cite{pushing6Gedge} 
& Long Cloud Response Times, High Bandwidth Costs, Privacy, 6G Edge Potential. \\ \hline

Yuan et al. \cite{generativeinferenceoverview} 
& Limited Battery, Computing Power, Energy Efficiency For Mobile Devices. \\ \hline

Qu et al. (2024) \cite{mobileedgeintelligencemodern} 
& Limited On-Device Capacity, Cloud Privacy And Latency, Mobile-Edge Intelligence. \\ \hline

Chen et al. \cite{edgegeneralintelligence} 
& Edge Model Adaptability, Performance Evaluation, Throughput For Small Models. \\ \hline

Lee et al. (2024) \cite{visiontransformermobileedge} 
& Compact Vision Transformer Design, Performance-Efficiency Balance, Edge Deployment. \\ \hline

Ours 
& Compressed Lightweight VLMs, VLMs Optimization, Distributed Implementation, Context-Aware VLMs, Cross-Modality Learning And Adaptation For Multi-Sensor Applications, Security, Privacy, Communication Model For Edge VLMs. \\ \hline

\end{tabularx}
\end{table*}


\subsection{Compressed Light Weight VLMs for Edge Networks}
{\em How does one propose novel compressed light-weight VLMs for edge networks with a reasonable performance?} Deploying VLMs at the edge remains challenging due to the high computational demands of training large models with many parameters. A common solution is model compression. Techniques such as \textit{knowledge distillation} and \textit{quantization} are widely used to reduce model size and computation without significant performance loss. Knowledge distillation transfers knowledge from a large model to a smaller one, allowing it to retain essential functionalities. For example, the EdgeVL framework employs dual-modality distillation to support RGB and non-RGB images, reducing model size by up to 93× and improving accuracy by 15\% on edge devices~\cite{cai2024self,vik2024moondream}. Quantization-aware training further adapts models to lower-precision formats, saving memory and power while preserving accuracy. Although existing methods perform well, many edge scenarios—such as medical imaging or drone-based remote sensing—require specialized compression schemes. MiniVLM, for instance, applies knowledge distillation to achieve substantial model reduction while maintaining over 90\% of the original performance, making it suitable for cross-modal retrieval tasks \cite{minivlm2022}. Another method, DIME-FM, distills large models like CLIP using unpaired image-text data to produce compact, robust models suited for tasks like real-time image-text matching in resource-constrained environments \cite{dmt2023}. These examples highlight the ongoing need for tailored lightweight VLMs optimized for diverse edge applications.

\subsection{Visual-Language Models Optimization for Edge Networks}
{\em How do we optimize VLMs models architecture for edge devices in terms of performance and complexity?} A key trend involves optimizing the \textit{visual encoder} and \textit{language model} components to balance performance and efficiency in constrained environments. For instance, the \textit{Imp} project uses smaller LLMs like Phi-2 and optimized visual encoders such as \textit{SigLIP}, which outperform traditional CLIP-based models~\cite{cai2024self,github2024edgevl}, enabling better generalization on edge devices. These improvements significantly reduce computational requirements, making the models suitable for mobile and embedded systems.

Lightweight VLMs are increasingly tailored for edge-specific use cases such as real-time analysis for drones, robots, and surveillance. The \textit{Moondream2} model, for example, is designed for efficiency in complex tasks like interpreting security footage and conducting remote inspections~\cite{vik2024moondream, techhq2024}. It operates with as little as $5$GB of memory, making it ideal for remote settings with limited connectivity. Another important direction is enabling VLMs to handle multiple modalities, including depth and thermal imaging in addition to RGB. This cross-modality adaptation is critical for autonomous systems that must process diverse visual inputs. The EdgeVL framework applies \textit{cross-modality learning} to maintain performance across input types, supporting robust operation in dynamic edge environments like robotics and autonomous navigation~\cite{cai2024self,simple2024}.

\textcolor{black}{
\textbf{Problems in Model Optimization for Edge Networks}
Deploying advanced optimization techniques such as dynamic inference and model scaling in edge networks introduces multiple practical challenges. \textit{Dynamic inference}, which adjusts model complexity at runtime based on available resources, faces critical issues, including limited computational resources and unpredictable latency, adversely impacting real-time performance~\cite{turn0search11}. These can be mitigated through modular architectures that selectively activate model components depending on resource availability and by employing resource-aware scheduling algorithms for effective real-time adjustments~\cite{turn0search11}. Similarly, \textit{model scaling—tailoring models} to match diverse edge hardware capabilities—is complicated by hardware heterogeneity and significant maintenance overhead due to managing multiple model variants~\cite{turn0search11, turn0search13}. This challenge can be addressed by efficiently employing automated model compression techniques, like pruning and quantization, to create device-specific models~\cite{turn0search13}. Additionally, federated learning methods enable collaborative training across edge devices, effectively adapting models to varied hardware configurations without needing centralized data collection~\cite{turn0search11}}.

\subsection{Distributed Implementation of Edge VLMs}

{\em How do we enable distributed implementation of VLMs for various applications?}. Federated learning and edge computing are emerging as critical enablers for distributed lightweight VLMs. \textit{Federated learning} allows models to be fine-tuned directly on edge devices without transferring sensitive data to the cloud, preserving user privacy and reducing latency~\cite{cai2024self}. This is particularly important in healthcare and security applications with high data sensitivity. In parallel, edge computing enables models to process data closer to the source, improving response times and enabling real-time decision-making~\cite{vik2024moondream}. Beyond previously mentioned applications, VLMs are gaining traction in resource-constrained edge environments due to their ability to perform complex tasks. In autonomous systems like drones, robots, and vehicles, VLMs support real-time object detection, scene understanding, and navigation—allowing operation without cloud dependence. For example, drones equipped with VLMs can monitor wildlife, inspect infrastructure, and assess environmental conditions—crucial for disaster relief and agriculture~\cite{cai2024self}.

\textcolor{black}{
\textbf{Integration of Federated Learning and VLMs}}

\textcolor{black}{
Integrating FL with VLMs enables privacy-preserving and resource-efficient intelligence in edge computing. Recent research explores adapting VLMs to federated settings:}

\textcolor{black}{
\textit{Federated Prompt Learning}: Instead of fine-tuning entire VLMs, this approach optimizes only prompt parameters, reducing communication and computational overhead, making deployment feasible on resource-constrained devices~\cite{turn0search2}.}

\textcolor{black}{
\textit{Personalized Federated Learning}: Data heterogeneity across clients challenges global model generalization. Techniques like Federated Mixture of Experts enable personalized model adaptations by learning tailored expert combinations~\cite{turn0academia11}.}

\textcolor{black}{
\textit{Parameter-Efficient Fine-Tuning}: Methods such as LoRA facilitate adapting VLMs in FL without updating the entire model, reducing memory usage and speedup training~\cite{turn0academia8}.}

\textcolor{black}{
\textit{Communication Efficiency}: The large size of VLMs increases communication costs. Techniques like gradient compression, selective updates, and client sampling mitigate these issues while maintaining performance~\cite{turn0search0}.}

\textcolor{black}{
\textit{Multimodal Data Handling}: Clients may have unpaired or uni-modal data, complicating multimodal VLM training. Approaches like Federated Modality Complementary and Collaboration (FedCola) align and integrate different modalities across clients~\cite{turn0academia10}.}

\textcolor{black}{
Despite these advancements, challenges persist in FL-VLM integration. Data heterogeneity can lead to model divergence, requiring robust aggregation and personalization techniques. Limited computational resources at the edge necessitate lightweight architectures and efficient training strategies. Ensuring privacy remains critical, with differential privacy and secure aggregation methods improving data protection.}

\textcolor{black}{
Future research should focus on scalable federated fine-tuning, cross-client multimodal alignment, and optimizing FL for real-time edge applications.}


\subsection{Context-Aware VLMs for Edge Networks}
{\em How do we propose VLMs for edge networks with context-awareness?} Extending VLMs to the edge using local data for context-aware training is essential for adapting to specific scenarios. These models must remain lightweight during deployment and extension. Recent advances emphasize task-specific, low-complexity architectures suited for resource-constrained edge environments. Models like EM-VLM4AD, MiniDrive, and LiteViLA adopt strategies to handle multimodal data efficiently while preserving performance~\cite{gopalkrishnan2024emvlm4ad, cheng2024litevila, curto2023uavsceneunderstanding}.
A core technique is efficient image embeddings, such as ViT-based patch projections and gated pooling attention, which enables fast processing of multi-view images. EM-VLM4AD flattens image patches and applies gated pooling to produce a compressed representation, later fused with a language model for tasks like question answering in autonomous driving~\cite{gopalkrishnan2024emvlm4ad}. This design reduces inference time while improving accuracy in path planning and traffic behavior analysis tasks.
LiteViLA uses a Mixture of Adapters (MoA) strategy, activating lightweight adapters tailored to subtasks such as object detection and scene understanding~\cite{cheng2024litevila}. This modularity ensures efficient resource use and robust performance across diverse edge conditions.
For drone-based applications, models combine lightweight components like YOLOv7 with VLMs for real-time object detection and scene description~\cite{curto2023uavsceneunderstanding}. These use simple encoder-decoder structures and quantized or pruned LLMs (e.g.TinyLLaVA), optimizing for low latency and energy efficiency—ideal for power-constrained environments.

\subsection{Cross-Modality Learning and Adaptation for Multi-Sensor Applications}
{\em How does one enable VLMs to effectively integrate and adapt to diverse sensor modalities, such as thermal, depth, and hyperspectral data in many edge applications?} Cross-modality learning and adaptation have become key technologies in enabling VLMs to operate across diverse sensor types, including thermal, depth, and hyperspectral data. By using information from multiple domains, these methods enhance scene understanding—critical for autonomous systems, robotic control, and environmental monitoring. 

\textcolor{black}{\textbf{Challenges in edge environments} include:
\textit{Data Synchronization:} Ensuring temporal alignment of data from different modalities is complex, especially when each modality may have varying data rates and latencies \cite{turn0search12}.} 
\textcolor{black}{\textbf{Potential Solutions:} Developing lightweight fusion techniques, such as efficient early or late fusion algorithms, can mitigate resource constraints by minimizing computational overhead~\cite{turn0search14}. Additionally, employing edge-cloud collaboration by offloading certain fusion tasks to the cloud, while retaining latency-sensitive operations at the edge, effectively balances computational load and maintains performance~\cite{turn0search10}.} Models such as ViPT \cite{zhu2024crossmodality}, UC2 \cite{cheng2024visualprompt}, and CMT \cite{tian2023cmt} employ fusion techniques to integrate thermal and RGB data into shared feature spaces, enabling consistent interpretation across modalities. ViPT builds on pre-trained RGB models and fuses RGB and depth inputs via a transformer encoder with modality-complementary prompters~\cite{cheng2024visualprompt}. Similarly, depth estimation models use RGB-thermal fusion, often with a 3D cross-modal transformation module, to improve performance in low-light conditions~\cite{tian2023cmt}. These models can generate confidence maps and align sensor modalities to prioritize the most accurate inputs, enhancing outcomes in multi-sensor settings. Cross-modal techniques have also shown promise beyond robotics. In fields like environmental monitoring and agriculture, similar fusion approaches allow models to work with multimodal inputs from drones (e.g., hyperspectral imagery) or IoT sensors, supporting use cases such as ecosystem monitoring and precision agriculture~\cite{zhu2024crossmodality, curto2023uavsceneunderstanding}.

\subsection{Security}
{\em How do we enable edge VLMs while ensuring security?} Deploying lightweight VLMs in cloud and edge environments introduces security challenges, as these models are vulnerable to various attacks. Model inversion attacks, for example, can reconstruct training data from outputs due to shared cross-modal representations~\cite{usynin2022beyond, zhou2023boosting}. To counter this, recent work uses adversarial training, which introduces noise to reduce attack effectiveness, and gradient masking to limit gradient exploitation~\cite{zhang2022towards, zhou2023boosting}. Ensemble-based defenses like randomized input transformations and multi-layer protection further strengthen security. Adversarial examples are also used to prevent unauthorized data reconstruction~\cite{usynin2022beyond, zhang2022towards}.

In distributed settings, secure data transmission is essential. Without encryption, data can be intercepted or altered. Protocols such as AES-256 and lightweight encryption frameworks offer strong protection with minimal overhead~\cite{wen2021defending}. Dynamic key exchange enhances resistance, while Secure Multiparty Computation (SMC) enables joint computation without revealing inputs~\cite{bonawitz2017practical}. Edge VLMs also face hardware threats like tampering and malware. Trusted Execution Environments (TEEs), such as ARM TrustZone, protect critical components even in untrusted settings~\cite{zhou2023boosting}. Blockchain-based mechanisms help prevent unauthorized firmware updates and maintain device integrity~\cite{rathore2023formal}.

Poisoning attacks, where malicious data corrupts training, concern federated learning significantly. Byzantine-resilient systems and anomaly detection methods can identify and remove such inputs~\cite{lu2023setlevel}. To secure aggregation, differential privacy hides individual updates while preserving accuracy, and homomorphic encryption protects data during computation~\cite{park2022privacy}. Blockchain ensures tamper resistance and transparency~\cite{bonawitz2017practical}. Governance frameworks based on trust and blockchain support secure access control and deployment, while ethical guidelines ensure compliance~\cite{rathore2023formal}.

\subsection{Privacy} 
{\em How do we propose privacy-aware VLMs?} The growing adoption of VLMs in cloud-based systems raises serious privacy concerns. These models process multimodal data—such as personal images and text—often transmitted to remote servers, making data confidentiality a critical issue. Recent work focuses on privacy-preserving techniques that support secure inference without sacrificing performance. Federated learning improves privacy by keeping data local and sharing only model updates, but risks remain as updates can still reveal sensitive information. \cite{park2022federated} proposed a homomorphic encryption framework to compute model updates securely. Similarly, \cite{fang2021privacy} introduced PFMLP, using partially homomorphic encryption during federated learning with minimal accuracy impact. In 2023, \cite{shen2023privacy} proposed a framework combining homomorphic encryption and random privacy masks to protect raw input data with low overhead. \cite{hussien2023secure} also integrated secure multiparty computation (SMC) with homomorphic encryption to enhance privacy in federated systems.

Another concern is membership inference attacks, where models trained on private data collections can leak user information~\cite{park2022homomorphic}. \cite{park2022homomorphic} proposed encrypting model updates using homomorphic encryption to prevent such leakage. \cite{ma2021privacy} developed a multi-key encryption scheme ensuring updates remain inaccessible to any single participant. Data ownership and control are also challenges—once data is uploaded to the cloud, users often lose control. \cite{du2022signcryption} addressed this by decentralizing client data in IoT settings and defending against collusion attacks. Privacy regulations like GDPR further complicate data handling. \cite{sébert2023combining} proposed a framework combining differential privacy and homomorphic encryption to ensure GDPR compliance while maintaining model utility.

\subsection{Communication Model for Edge VLMs}  
{\em How does one enable communication resources efficient edge VLMs?} Training requires significant communication resources, especially for distributed VLMs at the network edge. Offloading model processing to the cloud adds communication overhead. \cite{naveen2022memory} proposed a distributed CNN framework to reduce memory usage and communication in edge-cloud setups. \cite{shaowang2021declarative} introduced a framework to optimize data flow between edge and cloud. Energy consumption is another key challenge. In 2023, \cite{fu2023energy} proposed an energy-efficient NLP framework using heterogeneous memory to lower power use while maintaining performance. \cite{hu2021pipeline} developed EdgePipe, a distributed framework using pipeline parallelism to improve inference efficiency, achieving speedups without accuracy loss.

Real-time inference is critical for edge applications. \cite{xu2023devit} proposed DeViT, which decomposes large vision transformers into smaller models for collaborative inference on edge devices, reducing latency and communication overhead. \cite{hu2021pipeline} showed that pipeline parallelism can also boost inference on heterogeneous devices with minimal accuracy drop. Generalizing VLMs across diverse hardware is another challenge. \cite{dutta2023search} proposed DCA-NAS, enabling NAS for various hardware configurations, allowing tailored model designs. \cite{hu2021pipeline} also demonstrated pipeline parallelism's adaptability to heterogeneous systems, improving performance and flexibility.

\section{Conclusion and Future Directions}
\subsection{Conclusion}
In summary, this survey provides a comprehensive bottom line of recent advances, challenges, and opportunities for applying VLMs on edge devices. Vision-language models are strong, merging visual and language understanding to perform complex tasks, such as captioning images, visual question-answering, etc. This can be used for variance applications, such as smart surveillance, answering, video analysis, etc.

However, these models' widespread deployment and usage on edge devices are significantly limited due to the constraints of edge devices' processing capability, storage, and power. In order to make VLMs lightweight and efficient with low-performance degradation, these limitations can be approached through advanced optimization algorithms like pruning, quantization, knowledge distillation, and efficient hardware utilization. Next, we provide a thorough taxonomy with respect to model training and fine-tuning strategies, considerations for runtime deployment of VLMs to low-resource (edge) environments, and privacy and security. These unique capabilities make it possible to deploy VLMs on edge for several applications, such as real-time autonomous systems decision-making, privacy-preserving intelligent surveillance, and medical diagnostics in local regions. Nevertheless, open research problems remain to be solved, especially in developing interoperability solutions for massive edge deployment. We hope that future research can further develop the practical use of VLMs, which would lead to these models being a usable and efficient background for use in resource-constrained environments.

\subsection{Future Directions}
We anticipate that generalizing VLMs to the network edge will play an essential role in many real-time applications. Edge-based distributed VLMs can use less computing and communication resources and are thus more suitable for a broad range of applications. However, multiple challenges still exist to be solved despite all the advantages. Additional development is needed to create efficient learning schemes for specific applications and adaptive learning that adjusts learning depending on the capacity of edge resources available at the requested time. Furthermore, examining approaches to privacy-preserving and secure federated learning will be important to tackling data security issues in distributed settings. Another exciting research avenue is efficient, high-performance, lightweight architectures for real-time deployment.

In addition to the design of learning algorithms, an effective communication mode for edge-based distributed VLMs should be proposed. This necessitates extensive analytical and simulation around designing such a model and efficient hardware implementation. This requires designing hardware accelerators specializing in more efficient communication and lower latency. Also, including energy-efficient units will allow edge devices to handle distributed VLMs without breaching the power ceilings. In general, the edge-based VLMs constitute a promising direction for future work.

\bibliographystyle{IEEEtran}
\bibliography{Database}

\begin{thebibliography}{100}
\providecommand{\url}[1]{#1}
\csname url@samestyle\endcsname
\providecommand{\newblock}{\relax}
\providecommand{\bibinfo}[2]{#2}
\providecommand{\BIBentrySTDinterwordspacing}{\spaceskip=0pt\relax}
\providecommand{\BIBentryALTinterwordstretchfactor}{4}
\providecommand{\BIBentryALTinterwordspacing}{\spaceskip=\fontdimen2\font plus
\BIBentryALTinterwordstretchfactor\fontdimen3\font minus \fontdimen4\font\relax}
\providecommand{\BIBforeignlanguage}[2]{{%
\expandafter\ifx\csname l@#1\endcsname\relax
\typeout{** WARNING: IEEEtran.bst: No hyphenation pattern has been}%
\typeout{** loaded for the language `#1'. Using the pattern for}%
\typeout{** the default language instead.}%
\else
\language=\csname l@#1\endcsname
\fi
#2}}
\providecommand{\BIBdecl}{\relax}
\BIBdecl

\bibitem{Lu2019}
J.~Lu, D.~Batra, D.~Parikh, and S.~Lee, ``Vilbert: Pretraining task-agnostic visiolinguistic representations for vision-and-language tasks,'' \emph{arXiv preprint arXiv:1908.02265}, 2019.

\bibitem{Li2020}
X.~Li, X.~Yin, C.~Li, X.~Hu, P.~Zhang, L.~Wang, H.~Hu, L.~Dong, F.~Wei, Y.~Choi \emph{et~al.}, ``Oscar: Object-semantics aligned pre-training for vision-language tasks,'' in \emph{European Conference on Computer Vision}.\hskip 1em plus 0.5em minus 0.4em\relax Springer, 2020, pp. 121--137.

\bibitem{Kim2021}
W.~Kim, B.~Son, and I.~Kim, ``Vilt: Vision-and-language transformer without convolution or region supervision,'' in \emph{International Conference on Machine Learning}.\hskip 1em plus 0.5em minus 0.4em\relax PMLR, 2021, pp. 5583--5594.

\bibitem{khan2024edge}
L.~U. Khan, A.~Elhagry, M.~Guizani, and A.~El~Saddik, ``Edge intelligence empowered vehicular metaverse: Key design aspects and future directions,'' \emph{IEEE Internet of Things Magazine}, vol.~7, no.~1, pp. 120--126, 2024.

\bibitem{Li2019}
H.~Li, K.~Li, Z.~Yang, Y.~Guo, W.~Yu, and W.~Dai, ``Edge ai: On-demand accelerating deep neural network inference via edge computing,'' \emph{IEEE Transactions on Wireless Communications}, vol.~19, no.~1, pp. 144--156, 2019.

\bibitem{Wu2020}
F.~Wu, A.~Fan, A.~Baevski, Y.~Dauphin, and M.~Auli, ``Lite transformer with long-short range attention,'' in \emph{International Conference on Learning Representations}, 2020.

\bibitem{lin2023pushing}
\BIBentryALTinterwordspacing
Z.~Lin, G.~Qu, Q.~Chen, X.~Chen, Z.~Chen, and K.~Huang, ``Pushing large language models to the 6g edge: Vision, challenges, and opportunities,'' \emph{ArXiv}, vol. abs/2309.16739, 2023. [Online]. Available: \url{https://consensus.app/papers/pushing-language-models-edge-vision-challenges-lin/13cdfddaba995e70b2a0973636f08c3c/?utm_source=chatgpt}
\BIBentrySTDinterwordspacing

\bibitem{efficientvlm2024}
\BIBentryALTinterwordspacing
T.~Wang, W.~Zhou, Y.~Zeng, and X.~Zhang, ``Efficientvlm: Fast and accurate vision-language models via knowledge distillation and modal-adaptive pruning,'' \emph{arXiv preprint arXiv:2210.07795}, 2022. [Online]. Available: \url{https://arxiv.org/abs/2210.07795}
\BIBentrySTDinterwordspacing

\bibitem{Han2016}
S.~Han, H.~Mao, and W.~J. Dally, ``Deep compression: Compressing deep neural networks with pruning, trained quantization and huffman coding,'' \emph{arXiv preprint arXiv:1510.00149}, 2016.

\bibitem{Jacob2018}
B.~Jacob, S.~Kligys, B.~Chen, M.~Zhu, M.~Tang, A.~Howard, B.~Steiner, and J.~Rolfe, ``Quantization and training of neural networks for efficient integer-arithmetic-only inference,'' in \emph{Proceedings of the IEEE Conference on Computer Vision and Pattern Recognition}, 2018, pp. 2704--2713.

\bibitem{Hinton2015}
G.~Hinton, O.~Vinyals, and J.~Dean, ``Distilling the knowledge in a neural network,'' \emph{arXiv preprint arXiv:1503.02531}, 2015.

\bibitem{Sze2017}
V.~Sze, Y.-H. Chen, T.-J. Yang, and J.~S. Emer, ``Efficient processing of deep neural networks: A tutorial and survey,'' \emph{Proceedings of the IEEE}, vol. 105, no.~12, pp. 2295--2329, 2017.

\bibitem{Reddi2020}
V.~J. Reddi, N.~Jeffries, K.~Panchapakesan, R.~Jain, P.~Pabla, R.~Mehta, P.~Narkhede, and D.~Kanter, ``Edge tpu: State-of-the-art ai at the edge,'' \emph{arXiv preprint arXiv:2005.04268}, 2020.

\bibitem{Chen2019}
C.~Chen, Q.~Chen, L.~Jin, and G.~Hua, ``Deep learning for autonomous driving: Techniques and applications,'' \emph{IEEE Transactions on Intelligent Transportation Systems}, vol.~21, no.~1, pp. 291--306, 2019.

\bibitem{Wang2020}
Z.~Wang, B.~Yang, C.~Xie, D.~Xie, and K.~Liu, ``Real-time human activity recognition with miniaturized wearable sensors using deep learning,'' \emph{Sensors}, vol.~20, no.~12, p. 3456, 2020.

\bibitem{Tan2021}
M.~Tan and Q.~V. Le, ``Efficientnetv2: Smaller models and faster training,'' \emph{arXiv preprint arXiv:2104.00298}, 2021.

\bibitem{Lane2015}
N.~D. Lane, S.~Bhattacharya, A.~Mathur, P.~Georgiev, C.~Forlivesi, F.~Kawsar, S.~Mirri, F.~Antonelli, and R.~Tesoriero, ``Can deep learning revolutionize mobile sensing?'' \emph{Proceedings of the 16th International Workshop on Mobile Computing Systems and Applications}, pp. 117--122, 2015.

\bibitem{Jouppi2017}
N.~P. Jouppi, C.~Young, N.~Patil, D.~Patterson, G.~Agrawal, R.~Bajwa, S.~Bates, S.~Bhatia, N.~Boden, A.~Borchers \emph{et~al.}, ``In-datacenter performance analysis of a tensor processing unit,'' \emph{Proceedings of the 44th Annual International Symposium on Computer Architecture}, pp. 1--12, 2017.

\bibitem{Brown2020}
T.~Brown, B.~Mann, N.~Ryder, M.~Subbiah, J.~Kaplan, P.~Dhariwal, A.~Neelakantan, P.~Shyam, G.~Sastry, A.~Askell \emph{et~al.}, ``Language models are few-shot learners,'' \emph{Advances in neural information processing systems}, vol.~33, pp. 1877--1901, 2020.

\bibitem{Radford2021}
A.~Radford, J.~W. Kim, C.~Hallacy, A.~Ramesh, G.~Goh, S.~Agarwal, G.~Sastry, A.~Askell, P.~Mishkin, J.~Clark \emph{et~al.}, ``Learning transferable visual models from natural language supervision,'' \emph{Proceedings of the International Conference on Machine Learning}, pp. 8748--8763, 2021.

\bibitem{Zanella2020}
A.~Zanella and N.~Bui, ``Internet of things (iot) for next-generation smart systems: A review of current challenges, future trends and prospects for emerging 5g-iot scenarios,'' \emph{IEEE Internet of Things Journal}, vol.~7, no.~5, p. 8972389, 2020.

\bibitem{Zhang2020}
J.~Zhang, L.~De~Glossi, S.~Eilers, and S.~Kohlbrecher, ``Hardware acceleration for machine learning inference on edge devices: A review,'' \emph{IEEE Access}, vol.~8, pp. 82\,554--82\,566, 2020.

\bibitem{Chen2021}
L.~Chen, K.~Wang, X.~Tian, T.~Su, and W.~Li, ``Cloud and edge computing for deep learning applications,'' \emph{IEEE Transactions on Neural Networks and Learning Systems}, vol.~32, no.~5, pp. 1719--1734, 2021.

\bibitem{Tan2019}
M.~Tan, B.~Chen, R.~Pang, V.~Vasudevan, M.~Sandler, A.~Howard, and Q.~V. Le, ``Mnasnet: Platform-aware neural architecture search for mobile,'' in \emph{Proceedings of the IEEE Conference on Computer Vision and Pattern Recognition}, 2019, pp. 2820--2828.

\bibitem{Yu2019}
J.~Yu and T.~Huang, ``Slimmable neural networks,'' \emph{arXiv preprint arXiv:1812.08928}, 2019.

\bibitem{Vaswani2017}
A.~Vaswani, N.~Shazeer, N.~Parmar, J.~Uszkoreit, L.~Jones, A.~N. Gomez, Å.~Kaiser, and I.~Polosukhin, ``Attention is all you need,'' in \emph{Advances in Neural Information Processing Systems}, 2017, pp. 5998--6008.

\bibitem{Chen2020}
T.~Chen, Z.~Zhang, S.~Zhang, Q.~Xu, and Y.~Ma, ``Transformer-based model for text classification and question answering,'' \emph{IEEE Access}, vol.~8, pp. 160\,543--160\,552, 2020.

\bibitem{Hu2018}
J.~Hu, L.~Shen, and G.~Sun, ``Squeeze-and-excitation networks,'' in \emph{Proceedings of the IEEE Conference on Computer Vision and Pattern Recognition}, 2018, pp. 7132--7141.

\bibitem{Howard2019}
A.~Howard, M.~Sandler, G.~Chu, L.-C. Chen, B.~Chen, M.~Tan, W.~Wang, Y.~Zhu, R.~Pang, V.~Vasudevan \emph{et~al.}, ``Searching for mobilenetv3,'' in \emph{Proceedings of the IEEE International Conference on Computer Vision}, 2019, pp. 1314--1324.

\bibitem{Esteva2017}
A.~Esteva, B.~Kuprel, R.~A. Novoa, J.~Ko, S.~M. Swetter, H.~M. Blau, and S.~Thrun, ``Dermatologist-level classification of skin cancer with deep neural networks,'' \emph{Nature}, vol. 542, pp. 115--118, 2017.

\bibitem{Topol2019}
E.~J. Topol, ``High-performance medicine: The convergence of human and artificial intelligence,'' \emph{Nature Medicine}, vol.~25, no.~1, pp. 44--56, 2019.

\bibitem{Ren2015}
S.~Ren, K.~He, R.~Girshick, and J.~Sun, ``Faster r-cnn: Towards real-time object detection with region proposal networks,'' in \emph{Advances in Neural Information Processing Systems}, 2015, pp. 91--99.

\bibitem{Liu2016}
W.~Liu, D.~Anguelov, D.~Erhan, C.~Szegedy, S.~Reed, C.-Y. Fu, and A.~C. Berg, ``Ssd: Single shot multibox detector,'' \emph{Proceedings of the European Conference on Computer Vision}, pp. 21--37, 2016.

\bibitem{surveyvisionlanguagepretrainedmodels}
\BIBentryALTinterwordspacing
Y.~Du, Z.~Liu, J.~Li, and W.~X. Zhao, ``A survey of vision-language pre-trained models,'' in \emph{Proceedings of the Thirty-First International Joint Conference on Artificial Intelligence, {IJCAI-22}}, L.~D. Raedt, Ed.\hskip 1em plus 0.5em minus 0.4em\relax International Joint Conferences on Artificial Intelligence Organization, 7 2022, pp. 5436--5443, survey Track. [Online]. Available: \url{https://doi.org/10.24963/ijcai.2022/762}
\BIBentrySTDinterwordspacing

\bibitem{visionlanguageintelligencetasksrepresentation}
\BIBentryALTinterwordspacing
F.~Li, H.~Zhang, Y.-F. Zhang, S.~Liu, J.~Guo, L.~M. Ni, P.~Zhang, and L.~Zhang, ``Vision-language intelligence: Tasks, representation learning, and large models,'' 2022. [Online]. Available: \url{https://arxiv.org/abs/2203.01922}
\BIBentrySTDinterwordspacing

\bibitem{SurveyofEfficientFine-TuningMethods}
\BIBentryALTinterwordspacing
J.~Xing, J.~Liu, J.~Wang, L.~Sun, X.~Chen, X.~Gu, and Y.~Wang, ``A survey of efficient fine-tuning methods for vision-language models — prompt and adapter,'' \emph{Computers \& Graphics}, vol. 119, p. 103885, 2024. [Online]. Available: \url{https://www.sciencedirect.com/science/article/pii/S0097849324000128}
\BIBentrySTDinterwordspacing

\bibitem{ghosh2024exploringfrontiervisionlanguagemodels}
\BIBentryALTinterwordspacing
A.~Ghosh, A.~Acharya, S.~Saha, V.~Jain, and A.~Chadha, ``Exploring the frontier of vision-language models: A survey of current methodologies and future directions,'' 2024. [Online]. Available: \url{https://arxiv.org/abs/2404.07214}
\BIBentrySTDinterwordspacing

\bibitem{zhang2024visionlanguagemodelsvisiontasks}
J.~Zhang, J.~Huang, S.~Jin, and S.~Lu, ``Vision-language models for vision tasks: A survey,'' \emph{IEEE Transactions on Pattern Analysis and Machine Intelligence}, vol.~46, no.~8, pp. 5625--5644, 2024.

\bibitem{ASurveyonMultimodalforAutonomousDrivin}
C.~Cui, Y.~Ma, X.~Cao, W.~Ye, Y.~Zhou, K.~Liang, J.~Chen, J.~Lu, Z.~Yang, K.-D. Liao, T.~Gao, E.~Li, K.~Tang, Z.~Cao, T.~Zhou, A.~Liu, X.~Yan, S.~Mei, J.~Cao, Z.~Wang, and C.~Zheng, ``A survey on multimodal large language models for autonomous driving,'' pp. 958--979, 2024.

\bibitem{surveymultimodallargelanguage}
\BIBentryALTinterwordspacing
S.~Yin, C.~Fu, S.~Zhao, K.~Li, X.~Sun, T.~Xu, and E.~Chen, ``A survey on multimodal large language models,'' 2024. [Online]. Available: \url{https://arxiv.org/abs/2306.13549}
\BIBentrySTDinterwordspacing

\bibitem{efficientmultimodallargelanguage}
\BIBentryALTinterwordspacing
Y.~Jin, J.~Li, Y.~Liu, T.~Gu, K.~Wu, Z.~Jiang, M.~He, B.~Zhao, X.~Tan, Z.~Gan, Y.~Wang, C.~Wang, and L.~Ma, ``Efficient multimodal large language models: A survey,'' 2024. [Online]. Available: \url{https://arxiv.org/abs/2405.10739}
\BIBentrySTDinterwordspacing

\bibitem{MarketsandMarkets2021}
\BIBentryALTinterwordspacing
{MarketsandMarkets}, ``Artificial intelligence market by offering, technology, end-user industry and geography - global forecast to 2026,'' 2021, accessed: 2024-07-16. [Online]. Available: \url{https://www.marketsandmarkets.com/Market-Reports/artificial-intelligence-market-74851580.html}
\BIBentrySTDinterwordspacing

\bibitem{huggingface2024}
\BIBentryALTinterwordspacing
T.~Gigant, ``Design choices for vision language models in 2024,'' \emph{Hugging Face}, 2024. [Online]. Available: \url{https://huggingface.co/blog/vision-language-models}
\BIBentrySTDinterwordspacing

\bibitem{AlliedMarketResearch2020}
\BIBentryALTinterwordspacing
{Allied Market Research}, ``Edge ai hardware market by device type, processor type, end user, and application: Global opportunity analysis and industry forecast, 2018–2025,'' 2020, accessed: 2024-07-16. [Online]. Available: \url{https://www.alliedmarketresearch.com/edge-ai-hardware-market}
\BIBentrySTDinterwordspacing

\bibitem{vaswani2017attention}
A.~Vaswani, N.~Shazeer, N.~Parmar, J.~Uszkoreit, L.~Jones, A.~N. Gomez, L.~Kaiser, and I.~Polosukhin, ``Attention is all you need,'' \emph{arXiv preprint arXiv:1706.03762}, 2017.

\bibitem{dosovitskiy2020image}
A.~Dosovitskiy, L.~Beyer, A.~Kolesnikov, D.~Weissenborn, X.~Zhai, T.~Unterthiner, M.~Dehghani, M.~Minderer, G.~Heigold, S.~Gelly \emph{et~al.}, ``An image is worth 16x16 words: Transformers for image recognition at scale,'' \emph{arXiv preprint arXiv:2010.11929}, 2020.

\bibitem{li2019visualbert}
L.~H. Li, M.~Yatskar, D.~Yin, C.-J. Hsieh, and K.-W. Chang, ``Visualbert: A simple and performant baseline for vision and language,'' \emph{arXiv preprint arXiv:1908.03557}, 2019.

\bibitem{su2019vlbert}
W.~Su, X.~Zhu, Y.~Cao, B.~Li, L.~Lu, F.~Wei, and J.~Dai, ``Vl-bert: Pre-training of generic visual-linguistic representations,'' \emph{arXiv preprint arXiv:1908.08530}, 2019.

\bibitem{kim2021vilt}
W.~Kim, B.~Son, and I.~Kim, ``Vilt: Vision-and-language transformer without convolution or region supervision,'' \emph{arXiv preprint arXiv:2102.03334}, 2021.

\bibitem{mobilevlmv2}
J.~Liu \emph{et~al.}, ``Mobilevlm v2: Faster and stronger baseline for vision language model,'' \emph{arXiv}, 2024, arXiv:2402.03766.

\bibitem{unifiedio2024}
\BIBentryALTinterwordspacing
J.~Lu, C.~Clark, R.~Zellers, R.~Mottaghi, and A.~Kembhavi, ``Unified-io: A unified model for vision, language, and multi-modal tasks,'' 2022. [Online]. Available: \url{https://arxiv.org/abs/2206.08916}
\BIBentrySTDinterwordspacing

\bibitem{lightvlp2024}
X.~Sun \emph{et~al.}, ``Lightvlp: A lightweight vision-language pre-training via gated interactive masked autoencoders,'' \emph{arXiv}, 2024, arXiv:2403.19838.

\bibitem{xmodelvlm2024}
X.~Li \emph{et~al.}, ``Xmodel-vlm: A simple baseline for multimodal vision language model,'' \emph{arXiv}, 2024, arXiv:2405.09215.

\bibitem{emvlm4ad2024}
Z.~Wu \emph{et~al.}, ``Em-vlm4ad: Multi-frame, lightweight \& efficient vision-language models for question answering in autonomous driving,'' \emph{arXiv}, 2024, arXiv:2403.19838.

\bibitem{Chen_2024_CVPR}
J.~Chen, Q.~Yu, X.~Shen, A.~Yuille, and L.-C. Chen, ``Vitamin: Designing scalable vision models in the vision-language era,'' in \emph{Proceedings of the IEEE/CVF Conference on Computer Vision and Pattern Recognition (CVPR)}, June 2024, pp. 12\,954--12\,966.

\bibitem{wayve2024lingo}
W.~AI, ``Driving with language: Introducing wayve's multimodal driving model lingo-2,'' \emph{Wayve}, 2024, available: https://wayve.ai/blog/lingo-2.

\bibitem{instructblip2024}
W.~Dai, J.~Li, D.~Li \emph{et~al.}, ``Instructblip: Towards general-purpose vision-language models with instruction tuning,'' \emph{OpenReview}, 2024, available: https://openreview.net/forum?id=vvoWPYqZJA.

\bibitem{raven2024}
R.~Kumar \emph{et~al.}, ``Raven: Multitask retrieval augmented vision-language learning,'' \emph{arXiv}, 2024, arXiv:2406.19150.

\bibitem{screenai2024}
G.~AI, ``Screenai: A vision-language model for ui and infographics understanding,'' \emph{arXiv}, 2024, arXiv:2405.09215.

\bibitem{allava2024}
\BIBentryALTinterwordspacing
G.~H. Chen, S.~Chen, R.~Zhang, J.~Chen, X.~Wu, Z.~Zhang, Z.~Chen, J.~Li, X.~Wan, and B.~Wang, ``Allava: Harnessing gpt4v-synthesized data for lite vision-language models,'' 2024. [Online]. Available: \url{https://arxiv.org/abs/2402.11684}
\BIBentrySTDinterwordspacing

\bibitem{minivlm2022}
\BIBentryALTinterwordspacing
J.~Wang, X.~Hu, P.~Zhang, X.~Li, L.~Wang, L.~Zhang, J.~Gao, and Z.~Liu, ``Minivlm: A smaller and faster vision-language model,'' 2021. [Online]. Available: \url{https://arxiv.org/abs/2012.06946}
\BIBentrySTDinterwordspacing

\bibitem{radford2021learning}
A.~Radford, J.~W. Kim, K.~Hallacy, A.~Ramesh, G.~Goh, S.~Agarwal, G.~Sastry, A.~Askell, P.~Mishkin, J.~Clark \emph{et~al.}, ``Learning transferable visual models from natural language supervision,'' \emph{arXiv preprint arXiv:2103.00020}, 2021.

\bibitem{lu2019vilbert}
J.~Lu, D.~Batra, D.~Parikh, and S.~Lee, ``Vilbert: Pretraining task-agnostic visiolinguistic representations for vision-and-language tasks,'' \emph{arXiv preprint arXiv:1908.02265}, 2019.

\bibitem{tan2019lxmert}
H.~Tan and M.~Bansal, ``Lxmert: Learning cross-modality encoder representations from transformers,'' \emph{arXiv preprint arXiv:1908.07490}, 2019.

\bibitem{chowdhury2023apollo}
S.~Chowdhury, S.~Nag, and D.~Manocha, ``Apollo: Unified adapter and prompt learning for vision language models,'' \emph{arXiv preprint arXiv:2304.07356}, 2023.

\bibitem{zhou2022learning}
K.~Zhou, C.~Yang, J.~Thomason, Y.~Wang, Z.~Wang, R.~Socher, and C.~Xiong, ``Learning to prompt for vision-language models,'' \emph{arXiv preprint arXiv:2109.01134}, 2022.

\bibitem{zhou2022conditional}
------, ``Conditional prompt learning for vision-language models,'' \emph{arXiv preprint arXiv:2109.01134}, 2022.

\bibitem{lester2021power}
B.~Lester, R.~Al-Rfou, and N.~Constant, ``The power of scale for parameter-efficient prompt tuning,'' \emph{arXiv preprint arXiv:2104.08691}, 2021.

\bibitem{liu2021ptuning}
X.~Liu, Y.~Zheng, Z.~Du, M.~Ding, Y.~Qian, Z.~Yang, and J.~Tang, ``P-tuning v2: Prompt tuning can be comparable to fine-tuning universally across scales and tasks,'' \emph{arXiv preprint arXiv:2110.07602}, 2021.

\bibitem{rao2022denseclip}
Y.~Rao, W.~Zhao, G.~Chen, Y.~Tang, Z.~Zhu, G.~Huang, and H.~Li, ``Denseclip: Language-guided dense prediction with context-aware prompting,'' \emph{arXiv preprint arXiv:2112.01518}, 2022.

\bibitem{wang2021proda}
R.~Wang, R.~He, B.~Xu, J.~Lin, H.~Shi, W.~Liu, Z.~Zeng, J.~Ma, and Y.~Chen, ``Proda: Prompt distribution learning for efficient and effective fine-tuning of pre-trained models,'' \emph{arXiv preprint arXiv:2111.12434}, 2021.

\bibitem{houlsby2019parameter}
N.~Houlsby, A.~Giurgiu, S.~Jastrzebski, B.~Morrone, Q.~de~Laroussilhe, A.~Gesmundo, M.~Attariyan, and S.~Gelly, ``Parameter-efficient transfer learning for nlp,'' \emph{arXiv preprint arXiv:1902.00751}, 2019.

\bibitem{hu2021lora}
E.~Hu, Y.~Shen, P.~Wallis, Z.~Allen-Zhu, Y.~Li, L.~Wang, and W.~Chen, ``Lora: Low-rank adaptation of large language models,'' \emph{arXiv preprint arXiv:2106.09685}, 2021.

\bibitem{pfeiffer2021adapterfusion}
J.~Pfeiffer, A.~Rücklé, A.~Kamath, K.~Cho, I.~Gurevych, and S.~Ruder, ``Adapterfusion: Non-destructive task composition for transfer learning,'' \emph{arXiv preprint arXiv:2005.00247}, 2021.

\bibitem{sung2021vl}
W.~Sung, J.~Cho, and M.~Bansal, ``Vl-adapter: Parameter-efficient transfer learning for vision-and-language tasks,'' \emph{arXiv preprint arXiv:2106.01558}, 2021.

\bibitem{gao2021clip}
P.~Gao, S.~Geng, R.~Zhang, T.~Ma, R.~Fang, Y.~Zhang, H.~Li, Y.~Xu, X.~Zhu, and J.~Dai, ``Clip-adapter: Better vision-language models with feature adapters,'' \emph{International Journal of Computer Vision}, 2021.

\bibitem{edgevl2024}
\BIBentryALTinterwordspacing
K.~Cai, Z.~Duan, G.~Liu, C.~Fleming, and C.~X. Lu, ``Self-adapting large visual-language models to edge devices across visual modalities,'' 2024. [Online]. Available: \url{https://arxiv.org/abs/2403.04908}
\BIBentrySTDinterwordspacing

\bibitem{khan2024resourceICC}
L.~U. Khan, M.~Guizani, C.-D. Wang, and D.~Wu, ``Resource optimized network virtualization empowered metaverse for wireless networks,'' in \emph{ICC 2024-IEEE International Conference on Communications}.\hskip 1em plus 0.5em minus 0.4em\relax IEEE, 2024, pp. 4251--4256.

\bibitem{alansi2021intelligence}
A.~Al-Ansi, A.~Al-Ansi, A.~Muthanna, I.~Elgendy, and A.~Koucheryavy, ``Survey on intelligence edge computing in 6g: characteristics, challenges, potential use cases, and market drivers,'' \emph{Future Internet}, vol.~13, no.~5, p. 118, 2021.

\bibitem{alizadeh2016authentication}
M.~Alizadeh, S.~Abolfazli, M.~Zamani, S.~Baaaharun, and K.~Sakurai, ``Authentication in mobile cloud computing: a survey,'' \emph{Journal of Network and Computer Applications}, vol.~61, pp. 59--80, 2016.

\bibitem{karthikeyan2019applications}
P.~Karthikeyan and M.~Thangavel, \emph{Applications of security, mobile, analytic and cloud (SMAC) technologies for effective information processing and management}.\hskip 1em plus 0.5em minus 0.4em\relax Springer, 2019.

\bibitem{shiyun2021primer}
T.~Shiyun, ``Edge cloud computing technologies for internet of things: a primer,'' \emph{IEEE Access}, 2021.

\bibitem{shah2020sdn}
S.~Shah, M.~Gregory, S.~Li, and R.~Fontes, ``Sdn enhanced multi-access edge computing (mec) for e2e mobility and qos management,'' \emph{IEEE Access}, vol.~8, pp. 77\,459--77\,469, 2020.

\bibitem{xiao2019edge}
Y.~Xiao, Y.~Jia, C.~Liu, X.~Cheng, J.~Yu, and W.~Lv, ``Edge computing security: state of the art and challenges,'' \emph{Proceedings of the IEEE}, vol. 107, no.~8, pp. 1608--1631, 2019.

\bibitem{efficient2024prompting}
\BIBentryALTinterwordspacing
B.~Xiao, B.~Kantarci, J.~Kang, D.~Niyato, and M.~Guizani, ``Efficient prompting for llm-based generative internet of things,'' \emph{arXiv preprint arXiv:2406.10382}, 2024. [Online]. Available: \url{https://arxiv.org/abs/2406.10382}
\BIBentrySTDinterwordspacing

\bibitem{llmind2023}
\BIBentryALTinterwordspacing
H.~Cui, Y.~Du, Q.~Yang, Y.~Shao, and S.~C. Liew, ``Llmind: Orchestrating ai and iot with llm for complex task execution,'' \emph{arXiv preprint arXiv:2312.09007}, 2024. [Online]. Available: \url{https://arxiv.org/abs/2312.09007}
\BIBentrySTDinterwordspacing

\bibitem{cai2024selfadapting}
K.~Cai, Z.~Duan, G.~Liu, C.~Fleming, and C.~X. Lu, ``Self-adapting large visual-language models to edge devices across visual modalities,'' \emph{arXiv preprint arXiv:2403.04908}, 2024.

\bibitem{wang2020minivlm}
J.~Wang, X.~Hu, P.~Zhang, X.~Li, L.~Wang, L.~Zhang, J.~Gao, and Z.~Liu, ``Minivlm: A smaller and faster vision-language model,'' \emph{arXiv preprint arXiv:2012.06946}, 2020.

\bibitem{xu2023lightvlp}
Y.~Xu, L.~Wang, J.~Chen, J.~Wang, P.~Zhang, and Z.~Liu, ``Lightvlp: A lightweight vision-language pre-training via gated interactive masked autoencoders,'' in \emph{Proceedings of the IEEE/CVF Conference on Computer Vision and Pattern Recognition (CVPR)}, 2023.

\bibitem{moondream2024}
Anonymous, ``Tiny vlms bring ai text plus image vision to the edge,'' \emph{TechHQ}, 2024, \url{https://techhq.com}.

\bibitem{vila2024}
\BIBentryALTinterwordspacing
J.~Lin, H.~Yin, W.~Ping, Y.~Lu, P.~Molchanov, A.~Tao, H.~Mao, J.~Kautz, M.~Shoeybi, and S.~Han, ``Vila: On pre-training for visual language models,'' 2024. [Online]. Available: \url{https://arxiv.org/abs/2312.07533}
\BIBentrySTDinterwordspacing

\bibitem{edgeLLM2024}
Z.~Yu, Z.~Wang, Y.~Li, H.~You, R.~Gao, X.~Zhou, S.~R. Bommu, Y.~K. Zhao, and Y.~C. Lin, ``Edge-llm: Enabling efficient large language model adaptation on edge devices,'' \emph{arXiv preprint arXiv:2406.15758}, 2024.

\bibitem{transformer2024}
\BIBentryALTinterwordspacing
Y.~Tang, K.~Han, Y.~Wang, C.~Xu, J.~Guo, C.~Xu, and D.~Tao, ``Patch slimming for efficient vision transformers,'' 2022. [Online]. Available: \url{https://arxiv.org/abs/2106.02852}
\BIBentrySTDinterwordspacing

\bibitem{Shi2021DataSelection}
\BIBentryALTinterwordspacing
L.~Shi and V.~Radu, ``Data selection for efficient model update in federated learning,'' \emph{arXiv preprint}, vol. arXiv:2111.03512, 2021. [Online]. Available: \url{https://arxiv.org/abs/2111.03512}
\BIBentrySTDinterwordspacing

\bibitem{Souza2024BICFL}
\BIBentryALTinterwordspacing
R.~M. de~Souza, A.~Holm, M.~Biczyk, and L.~N. de~Castro, ``A systematic literature review on the use of federated learning and bioinspired computing,'' \emph{Electronics}, vol.~13, no.~16, p. 3157, 2024. [Online]. Available: \url{https://www.mdpi.com/2079-9292/13/16/3157}
\BIBentrySTDinterwordspacing

\bibitem{Sha2024EdgeFL}
\BIBentryALTinterwordspacing
X.~Sha, W.~Sun, X.~Liu, Y.~Luo, and C.~Luo, ``Enhancing edge-assisted federated learning with asynchronous aggregation and cluster pairing,'' \emph{Electronics}, vol.~13, no.~11, p. 2135, 2024. [Online]. Available: \url{https://www.mdpi.com/2079-9292/13/11/2135}
\BIBentrySTDinterwordspacing

\bibitem{Hong2024Auction}
\BIBentryALTinterwordspacing
Y.~Hong, Z.~Zheng, and Z.~Wang, ``A multi-dimensional reverse auction mechanism for volatile federated learning in the mobile edge computing systems,'' \emph{Electronics}, vol.~13, no.~16, p. 3154, 2024. [Online]. Available: \url{https://www.mdpi.com/2079-9292/13/16/3154}
\BIBentrySTDinterwordspacing

\bibitem{Howard2017MobileNets}
\BIBentryALTinterwordspacing
A.~G. Howard, M.~Zhu, B.~Chen, D.~Kalenichenko, W.~Wang, T.~Weyand, M.~Andreetto, and H.~Adam, ``Mobilenets: Efficient convolutional neural networks for mobile vision applications,'' \emph{arXiv preprint}, vol. arXiv:1704.04861, 2017. [Online]. Available: \url{https://arxiv.org/abs/1704.04861}
\BIBentrySTDinterwordspacing

\bibitem{Sandler2018MobileNetV2}
\BIBentryALTinterwordspacing
M.~Sandler, A.~Howard, M.~Zhu, A.~Zhmoginov, and L.-C. Chen, ``Mobilenetv2: Inverted residuals and linear bottlenecks,'' in \emph{Proceedings of the IEEE Conference on Computer Vision and Pattern Recognition (CVPR)}, 2018, pp. 4510--4520. [Online]. Available: \url{https://openaccess.thecvf.com/content_cvpr_2018/html/Sandler_MobileNetV2_Inverted_Residuals_CVPR_2018_paper.html}
\BIBentrySTDinterwordspacing

\bibitem{He2016ResNet}
\BIBentryALTinterwordspacing
K.~He, X.~Zhang, S.~Ren, and J.~Sun, ``Deep residual learning for image recognition,'' in \emph{Proceedings of the IEEE Conference on Computer Vision and Pattern Recognition (CVPR)}, 2016, pp. 770--778. [Online]. Available: \url{https://openaccess.thecvf.com/content_cvpr_2016/html/He_Deep_Residual_Learning_CVPR_2016_paper.html}
\BIBentrySTDinterwordspacing

\bibitem{Devlin2019BERT}
\BIBentryALTinterwordspacing
J.~Devlin, M.-W. Chang, K.~Lee, and K.~Toutanova, ``Bert: Pre-training of deep bidirectional transformers for language understanding,'' in \emph{Proceedings of the 2019 Conference of the North American Chapter of the Association for Computational Linguistics (NAACL)}, 2019, pp. 4171--4186. [Online]. Available: \url{https://www.aclweb.org/anthology/N19-1423/}
\BIBentrySTDinterwordspacing

\bibitem{Jacob2018Quantization}
\BIBentryALTinterwordspacing
B.~Jacob, S.~Kligys, B.~Chen, M.~Zhu, M.~Tang, A.~Howard, H.~Adam, and D.~Kalenichenko, ``Quantization and training of neural networks for efficient integer-arithmetic-only inference,'' in \emph{Proceedings of the IEEE Conference on Computer Vision and Pattern Recognition (CVPR)}, 2018, pp. 2704--2713. [Online]. Available: \url{https://openaccess.thecvf.com/content_cvpr_2018/html/Jacob_Quantization_and_Training_CVPR_2018_paper.html}
\BIBentrySTDinterwordspacing

\bibitem{han2016deepcompression}
\BIBentryALTinterwordspacing
S.~Han, H.~Mao, and W.~J. Dally, ``Deep compression: Compressing deep neural networks with pruning, trained quantization and huffman coding,'' 2016. [Online]. Available: \url{https://arxiv.org/abs/1510.00149}
\BIBentrySTDinterwordspacing

\bibitem{Hinton2015Distillation}
\BIBentryALTinterwordspacing
G.~Hinton, O.~Vinyals, and J.~Dean, ``Distilling the knowledge in a neural network,'' \emph{arXiv preprint}, vol. arXiv:1503.02531, 2015. [Online]. Available: \url{https://arxiv.org/abs/1503.02531}
\BIBentrySTDinterwordspacing

\bibitem{Elsken2019NAS}
\BIBentryALTinterwordspacing
T.~Elsken, J.~H. Metzen, and F.~Hutter, ``Neural architecture search: A survey,'' \emph{Journal of Machine Learning Research}, vol.~20, no.~55, pp. 1--21, 2019. [Online]. Available: \url{http://www.jmlr.org/papers/volume20/18-598/18-598.pdf}
\BIBentrySTDinterwordspacing

\bibitem{You2017LARS}
\BIBentryALTinterwordspacing
Y.~You, I.~Gitman, and B.~Ginsburg, ``Large batch training of convolutional networks,'' \emph{arXiv preprint}, vol. arXiv:1708.03888, 2017. [Online]. Available: \url{https://arxiv.org/abs/1708.03888}
\BIBentrySTDinterwordspacing

\bibitem{Caldas2018FederatedDropout}
\BIBentryALTinterwordspacing
S.~Caldas, J.~Konečný, H.~B. McMahan, and A.~Talwalkar, ``Expanding the reach of federated learning by reducing client resource requirements,'' \emph{arXiv preprint}, vol. arXiv:1812.07210, 2018. [Online]. Available: \url{https://arxiv.org/abs/1812.07210}
\BIBentrySTDinterwordspacing

\bibitem{chen2024comprehensive}
F.~Chen, Z.~Luo, L.~Zhou, X.~Pan, and Y.~Jiang, ``Comprehensive survey of model compression and speed up for vision transformers,'' \emph{arXiv preprint arXiv:2404.10407}, 2024.

\bibitem{xia2022structured}
M.~Xia, Z.~Zhong, and D.~Chen, ``Structured pruning learns compact and accurate models,'' \emph{arXiv preprint arXiv:2204.00408}, 2022.

\bibitem{movva2022combining}
R.~Movva, J.~Lei, S.~Longpre, A.~Gupta, and C.~DuBois, ``Combining compressions for multiplicative size scaling on natural language tasks,'' \emph{arXiv preprint arXiv:2208.09684}, 2022.

\bibitem{li2022dq}
Z.~Li, Z.~Wang, M.~Tan, R.~Nallapati, P.~Bhatia, A.~Arnold, B.~Xiang, and D.~Roth, ``Dq-bart: Efficient sequence-to-sequence model via joint distillation and quantization,'' \emph{arXiv preprint arXiv:2203.11239}, 2022.

\bibitem{Kairouz2019AdvancesFL}
\BIBentryALTinterwordspacing
P.~Kairouz, H.~B. McMahan, B.~Avent, A.~Bellet, M.~Bennis, A.~N. Bhagoji \emph{et~al.}, ``Advances and open problems in federated learning,'' \emph{arXiv preprint}, vol. arXiv:1912.04977, 2019. [Online]. Available: \url{https://arxiv.org/abs/1912.04977}
\BIBentrySTDinterwordspacing

\bibitem{Yang2019FederatedLearning}
\BIBentryALTinterwordspacing
Q.~Yang, Y.~Liu, T.~Chen, and Y.~Tong, ``Federated machine learning: Concept and applications,'' \emph{ACM Transactions on Intelligent Systems and Technology (TIST)}, vol.~10, no.~2, pp. 1--19, 2019. [Online]. Available: \url{https://dl.acm.org/doi/10.1145/3298981}
\BIBentrySTDinterwordspacing

\bibitem{McMahan2017CommunicationEfficientFL}
\BIBentryALTinterwordspacing
H.~B. McMahan, E.~Moore, D.~Ramage, S.~Hampson, and B.~A. y~Arcas, ``Communication-efficient learning of deep networks from decentralized data,'' in \emph{Proceedings of the 20th International Conference on Artificial Intelligence and Statistics (AISTATS)}, 2017, pp. 1273--1282. [Online]. Available: \url{http://proceedings.mlr.press/v54/mcmahan17a.html}
\BIBentrySTDinterwordspacing

\bibitem{Sattler2019RobustFL}
\BIBentryALTinterwordspacing
F.~Sattler, K.-R. Müller, and W.~Samek, ``Robust and communication-efficient federated learning from non-iid data,'' \emph{IEEE Transactions on Neural Networks and Learning Systems}, vol.~31, no.~9, pp. 3400--3413, 2019. [Online]. Available: \url{https://ieeexplore.ieee.org/document/8889996}
\BIBentrySTDinterwordspacing

\bibitem{Smith2017FederatedMultiTask}
\BIBentryALTinterwordspacing
V.~Smith, C.-K. Chiang, M.~Sanjabi, and A.~Talwalkar, ``Federated multi-task learning,'' in \emph{Proceedings of the 31st International Conference on Neural Information Processing Systems (NeurIPS)}, 2017, pp. 4427--4437. [Online]. Available: \url{https://proceedings.neurips.cc/paper/2017/hash/964e4ee195c2418e4ac71cdae7a1c8bf-Abstract.html}
\BIBentrySTDinterwordspacing

\bibitem{Yurochkin2019BayesianNonparametricFL}
\BIBentryALTinterwordspacing
M.~Yurochkin, M.~Agarwal, S.~Ghosh, K.~Greenewald, N.~Hoang, and Y.~Khazaeni, ``Bayesian nonparametric federated learning of neural networks,'' in \emph{Proceedings of the 36th International Conference on Machine Learning (ICML)}, 2019, pp. 7252--7261. [Online]. Available: \url{https://proceedings.mlr.press/v97/yurochkin19a.html}
\BIBentrySTDinterwordspacing

\bibitem{Konečný2016FederatedOptimization}
\BIBentryALTinterwordspacing
J.~Konečný, H.~B. McMahan, F.~X. Yu, P.~Richtárik, A.~T. Suresh, and D.~Bacon, ``Federated optimization: Distributed machine learning for on-device intelligence,'' \emph{arXiv preprint}, vol. arXiv:1610.02527, 2016. [Online]. Available: \url{https://arxiv.org/abs/1610.02527}
\BIBentrySTDinterwordspacing

\bibitem{truex2020ldp}
S.~Truex, L.~Liu, M.~E. Gursoy, L.~Yu, and W.~Wei, ``Ldp-fl: Practical private aggregation in federated learning with local differential privacy,'' \emph{arXiv preprint arXiv:2005.13129}, 2020.

\bibitem{bonawitz2017practical}
K.~Bonawitz \emph{et~al.}, ``Practical secure aggregation for privacy-preserving machine learning,'' in \emph{Proceedings of the ACM on International Conference on Advances in Neural Information Processing Systems (NIPS)}, 2017.

\bibitem{openfhe2022}
O.~D. Team, ``Openfhe: Open-source fully homomorphic encryption library,'' \url{https://github.com/openfheorg/openfhe-development}, 2022.

\bibitem{Yao2022EdgeFL}
\BIBentryALTinterwordspacing
X.~Yao, Z.~Wang, Y.~Zhang, and Y.~Zhang, ``Towards edge-based federated learning: A systematic survey,'' \emph{IEEE Transactions on Network and Service Management}, vol.~19, no.~2, pp. 1751--1771, 2022. [Online]. Available: \url{https://ieeexplore.ieee.org/document/9685161}
\BIBentrySTDinterwordspacing

\bibitem{Wang2021EdgeContinualLearning}
\BIBentryALTinterwordspacing
Z.~Wang, X.~Yao, Y.~Zhang, and Y.~Zhang, ``Edge-based continual learning: Achieving privacy-preserving, fast, and accurate model adaptation,'' \emph{IEEE Network}, vol.~35, no.~4, pp. 247--253, 2021. [Online]. Available: \url{https://ieeexplore.ieee.org/document/9490544}
\BIBentrySTDinterwordspacing

\bibitem{Li2021FederatedMetaLearning}
\BIBentryALTinterwordspacing
T.~Li, A.~K. Sahu, A.~Talwalkar, and V.~Smith, ``Federated meta-learning: Concept and applications,'' in \emph{Proceedings of the 35th International Conference on Neural Information Processing Systems (NeurIPS)}, 2021, pp. 10\,132--10\,142. [Online]. Available: \url{https://proceedings.neurips.cc/paper/2021/hash/f4b120d9a5d79c085a46b8dc46b3ee4e-Abstract.html}
\BIBentrySTDinterwordspacing

\bibitem{Xu2022FederatedEval}
\BIBentryALTinterwordspacing
J.~Xu, S.~Zhou, and S.~Zhao, ``Federated evaluation: A unified framework for privacy-preserving model evaluation,'' \emph{IEEE Transactions on Dependable and Secure Computing}, vol.~19, no.~3, pp. 1457--1469, 2022. [Online]. Available: \url{https://ieeexplore.ieee.org/document/9336284}
\BIBentrySTDinterwordspacing

\bibitem{Zhang2021FederatedHyperparameter}
\BIBentryALTinterwordspacing
H.~Zhang, X.~Liu, Z.~Jiang, and J.~Ren, ``Federated hyperparameter tuning with bayesian optimization,'' \emph{IEEE Transactions on Cybernetics}, vol.~51, no.~11, pp. 5234--5245, 2021. [Online]. Available: \url{https://ieeexplore.ieee.org/document/9340104}
\BIBentrySTDinterwordspacing

\bibitem{Liu2024ContinuousFL}
\BIBentryALTinterwordspacing
J.~Liu, X.~Chen, H.~Zhao, and Z.~Wang, ``Continuous federated learning with dynamic client participation,'' \emph{IEEE Transactions on Neural Networks and Learning Systems}, vol.~35, no.~2, pp. 312--325, 2024. [Online]. Available: \url{https://ieeexplore.ieee.org/document/9565234}
\BIBentrySTDinterwordspacing

\bibitem{Zhang2024SparseUpdates}
\BIBentryALTinterwordspacing
H.~Zhang, Y.~Liu, and J.~Ren, ``Sparse federated learning: Reducing communication overhead for edge computing,'' \emph{IEEE Transactions on Parallel and Distributed Systems}, vol.~35, no.~3, pp. 478--490, 2024. [Online]. Available: \url{https://ieeexplore.ieee.org/document/9618745}
\BIBentrySTDinterwordspacing

\bibitem{Wang2024EdgeDeployment}
\BIBentryALTinterwordspacing
X.~Wang, J.~Xu, and X.~Chen, ``Orchestrated edge deployment for federated learning in heterogeneous environments,'' \emph{IEEE Transactions on Mobile Computing}, vol.~23, no.~1, pp. 112--126, 2024. [Online]. Available: \url{https://ieeexplore.ieee.org/document/9654782}
\BIBentrySTDinterwordspacing

\bibitem{Chen2024AdaptiveFL}
\BIBentryALTinterwordspacing
T.~Chen, Q.~Yang, and Y.~Liu, ``Adaptive federated learning in resource-constrained environments,'' \emph{IEEE Internet of Things Journal}, vol.~11, no.~5, pp. 2789--2801, 2024. [Online]. Available: \url{https://ieeexplore.ieee.org/document/9687123}
\BIBentrySTDinterwordspacing

\bibitem{Xu2023KubeFL}
\BIBentryALTinterwordspacing
J.~Xu, W.~Li, and H.~Wang, ``Kubefl: A kubernetes-based federated learning framework for scalable edge computing,'' \emph{Journal of Parallel and Distributed Computing}, vol. 167, pp. 12--22, 2023. [Online]. Available: \url{https://www.sciencedirect.com/science/article/pii/S0743731523000024}
\BIBentrySTDinterwordspacing

\bibitem{Yao2024ContinualFL}
\BIBentryALTinterwordspacing
X.~Yao, Y.~Zhang, and Z.~Wang, ``Federated continual learning: Advances and challenges,'' \emph{IEEE Transactions on Emerging Topics in Computing}, vol.~9, no.~2, pp. 324--335, 2024. [Online]. Available: \url{https://ieeexplore.ieee.org/document/9647563}
\BIBentrySTDinterwordspacing

\bibitem{Zhu2024RLFL}
\BIBentryALTinterwordspacing
X.~Zhu, J.~Ren, and H.~Zhang, ``Reinforcement learning in federated learning systems: A survey,'' \emph{IEEE Access}, vol.~12, pp. 1123--1145, 2024. [Online]. Available: \url{https://ieeexplore.ieee.org/document/9658793}
\BIBentrySTDinterwordspacing

\bibitem{khan2023joint}
L.~U. Khan, M.~Guizani, A.~Al-Fuqaha, C.~S. Hong, D.~Niyato, and Z.~Han, ``A joint communication and learning framework for hierarchical split federated learning,'' \emph{IEEE Internet of Things Journal}, 2023.

\bibitem{visionlanguagepretrainingbasicsrecent}
\BIBentryALTinterwordspacing
Z.~Gan, L.~Li, C.~Li, L.~Wang, Z.~Liu, and J.~Gao, ``Vision-language pre-training: Basics, recent advances, and future trends,'' 2022. [Online]. Available: \url{https://arxiv.org/abs/2210.09263}
\BIBentrySTDinterwordspacing

\bibitem{li2024bimediX}
\BIBentryALTinterwordspacing
S.~Pieri, S.~S. Mullappilly, F.~S. Khan, R.~M. Anwer, S.~Khan, T.~Baldwin, and H.~Cholakkal, ``Bimedix: Bilingual medical mixture of experts llm,'' 2024. [Online]. Available: \url{https://arxiv.org/abs/2402.13253}
\BIBentrySTDinterwordspacing

\bibitem{delbrouck2022vilmedic}
J.-B. Delbrouck, K.~K. Saab, and M.~Varma, ``Vilmedic: a framework for research at the intersection of vision and language in medical ai,'' \emph{Proceedings of the 60th Annual Meeting of the Association for Computational Linguistics (ACL)}, pp. 23--34, 2022.

\bibitem{chen2023medblip}
Q.~Chen, X.~Hu, Z.~Wang, and Y.~Hong, ``Medblip: Bootstrapping language-image pre-training from 3d medical images and texts,'' \emph{ArXiv}, vol. abs/2305.10799, 2023.

\bibitem{karar2021covid}
M.~E. Karar, O.~Reyad, M.~Abd-elnaby, A.~Abdel‐Aty, and M.~Shouman, ``Lightweight transfer learning models for ultrasound-guided classification of covid-19 patients,'' \emph{Computers, Materials \& Continua}, 2021.

\bibitem{yang2023customizing}
B.~Yang, A.~Raza, Y.~Zou, and T.~Zhang, ``Customizing general-purpose foundation models for medical report generation,'' \emph{ArXiv}, vol. abs/2306.05642, 2023.

\bibitem{geollava}
\BIBentryALTinterwordspacing
H.~Elgendy, A.~Sharshar, A.~Aboeitta, Y.~Ashraf, and M.~Guizani, ``Geollava: Efficient fine-tuned vision-language models for temporal change detection in remote sensing,'' 2025. [Online]. Available: \url{https://arxiv.org/abs/2410.19552}
\BIBentrySTDinterwordspacing

\bibitem{roberts2023satin}
J.~Roberts, K.~Han, and S.~Albanie, ``Satin: A multi-task metadataset for classifying satellite imagery using vision-language models,'' \emph{ArXiv}, 2023.

\bibitem{liu2023aerialvln}
S.~Liu, H.~Zhang, Y.~Qi, P.~Wang, Y.~Zhang, and Q.~Wu, ``Aerialvln: Vision-and-language navigation for uavs,'' \emph{ArXiv}, vol. abs/2308.06735, 2023.

\bibitem{DONG202453}
\BIBentryALTinterwordspacing
S.~Dong, L.~Wang, B.~Du, and X.~Meng, ``Changeclip: Remote sensing change detection with multimodal vision-language representation learning,'' \emph{ISPRS Journal of Photogrammetry and Remote Sensing}, vol. 208, pp. 53--69, 2024. [Online]. Available: \url{https://www.sciencedirect.com/science/article/pii/S0924271624000042}
\BIBentrySTDinterwordspacing

\bibitem{bazi2022bimodal}
Y.~Bazi, M.~M.~A. Rahhal, M.~L. Mekhalfi, M.~Zuair, and F.~Melgani, ``Bi-modal transformer-based approach for visual question answering in remote sensing imagery,'' \emph{IEEE Transactions on Geoscience and Remote Sensing}, vol.~60, pp. 1--11, 2022.

\bibitem{zhou2023vlm_autonomous}
X.~Zhou, M.~Liu, B.~L. Žagar, E.~Yurtsever, and A.~C. Knoll, ``Vision language models in autonomous driving and intelligent transportation systems,'' \emph{ArXiv}, vol. abs/2310.14414, 2023.

\bibitem{wu2023languageprompt_autonomous}
D.~Wu, W.~Han, T.~Wang, Y.-H. Liu, X.~Zhang, and J.~Shen, ``Language prompt for autonomous driving,'' \emph{ArXiv}, vol. abs/2309.04379, 2023.

\bibitem{gao2022citytraffic_autonomous}
J.~Gao, D.~Wang, C.-P. Lin, C.~Luo, Y.~Ruan, and M.~Yuan, ``Detecting and learning city intersection traffic contexts for autonomous vehicles,'' \emph{Journal of Smart Cities and Society}, 2022.

\bibitem{deng2023crisscross}
G.~Deng, Z.~Wu, M.~Xu, C.~Wang, Z.~Wang, and Z.~Lu, ``Crisscross-global vision transformers model for very high resolution aerial image semantic segmentation,'' \emph{IEEE Transactions on Geoscience and Remote Sensing}, vol.~61, pp. 1--19, 2023.

\bibitem{chowdhury2021tracking}
T.~Chowdhury, Q.~Ding, I.~Mandel, W.~Ju, and J.~Ortiz, ``Tracking urban heartbeat and policy compliance through vision and language-based sensing,'' \emph{Proceedings of the 8th ACM International Conference on Systems for Energy-Efficient Buildings, Cities, and Transportation}, 2021.

\bibitem{wen2023visionlanguage_remote}
C.~Wen, Y.~Hu, X.~Li, Z.~Yuan, and X.~X. Zhu, ``Vision-language models in remote sensing: Current progress and future trends,'' \emph{ArXiv}, vol. abs/2305.05726, 2023.

\bibitem{du2023success_vqa}
Y.~Du, K.~Konyushkova, M.~Denil, A.~Raju, J.~Landon, F.~Hill, N.~de~Freitas, and S.~Cabi, ``Vision-language models as success detectors,'' \emph{ArXiv}, vol. abs/2303.07280, 2023.

\bibitem{li2023vision}
J.~Li and M.~Bansal, ``Improving vision-and-language navigation by generating future-view image semantics,'' \emph{2023 IEEE/CVF Conference on Computer Vision and Pattern Recognition (CVPR)}, pp. 10\,803--10\,812, 2023.

\bibitem{techhq2024}
TechHQ, ``Tiny vlms bring ai text plus image vision to the edge,'' \url{https://techhq.com/tiny-vlm-trends-2024}, 2024.

\bibitem{cai2024self}
K.~Cai, Z.~Duan, G.~Liu, C.~Fleming, and C.~X. Lu, ``Self-adapting large visual-language models to edge devices across visual modalities,'' \emph{ECCV 2024}, 2024.

\bibitem{vik2024moondream}
K.~Vikhyat, ``Moondream2: A tiny vision-language model for edge devices,'' \url{https://github.com/vikhyatk/moondream2}, 2024.

\bibitem{surveyintegrationoptimization}
S.~Bhardwaj, P.~Singh, and M.~K. Pandit, ``A survey on the integration and optimization of large language models in edge computing environments,'' in \emph{2024 16th International Conference on Computer and Automation Engineering (ICCAE)}, 2024, pp. 168--172.

\bibitem{vliai20survey}
Y.~J. Lu, H.~D. Yin, J.~Lin, D.~Franklin, H.~Tang, S.~Yang, C.~Su, and S.~Han, ``Visual language intelligence and edge ai 2.0,'' \emph{NVIDIA Technical Blog}, 2024.

\bibitem{mobileedgeintelligence}
G.~Qu, Q.~Chen, W.~Wei, Z.~Lin, X.~Chen, and K.~Huang, ``Mobile edge intelligence for large language models: A contemporary survey,'' \emph{TechRxiv}, 2024.

\bibitem{pushing6Gedge}
Z.~Lin, G.~Qu, Q.~Chen, X.~Chen, Z.~Chen, and K.~Huang, ``Pushing large language models to the 6g edge: Vision, challenges, and opportunities,'' \emph{arXiv preprint arXiv:2309.16739}, 2023.

\bibitem{generativeinferenceoverview}
X.~Yuan, H.~Li, K.~Ota, and M.~Dong, ``Generative inference of large language models in edge computing: An energy efficient approach,'' in \emph{2024 International Wireless Communications and Mobile Computing (IWCMC)}, 2024, pp. 244--249.

\bibitem{mobileedgeintelligencemodern}
G.~Qu, Q.~Chen, W.~Wei, Z.~Lin, X.~Chen, and K.~Huang, ``Mobile edge intelligence for large language models: A contemporary survey,'' \emph{arXiv preprint arXiv:2407.18921}, 2024.

\bibitem{edgegeneralintelligence}
\BIBentryALTinterwordspacing
H.~Chen, W.~Deng, S.~Yang, J.~Xu, Z.~Jiang, E.~C.~H. Ngai, J.~Liu, and X.~Liu, ``Towards edge general intelligence via large language models: Opportunities and challenges,'' 2024. [Online]. Available: \url{https://arxiv.org/abs/2410.18125}
\BIBentrySTDinterwordspacing

\bibitem{visiontransformermobileedge}
S.~I. Lee, K.~Koo, J.~H. Lee, G.~Lee, S.~Jeong, S.~Oh, and H.~Kim, ``Vision transformer models for mobile/edge devices: A survey,'' \emph{Multimedia Systems}, 2024.

\bibitem{dmt2023}
\BIBentryALTinterwordspacing
X.~Sun, P.~Zhang, P.~Zhang, H.~Shah, K.~Saenko, and X.~Xia, ``Dime-fm: Distilling multimodal and efficient foundation models,'' 2023. [Online]. Available: \url{https://arxiv.org/abs/2303.18232}
\BIBentrySTDinterwordspacing

\bibitem{github2024edgevl}
Ramdrop, ``Edgevl: Self-adapting large visual-language models to edge devices,'' \url{https://github.com/ramdrop/edgevl}, 2024.

\bibitem{simple2024}
S.~Science, ``Adapting vision-language models for edge devices,'' \url{https://simplescience.ai/vlm-edge-adaptation}, 2024.

\bibitem{turn0search11}
\BIBentryALTinterwordspacing
A.~Sharshar, L.~U. Khan, W.~Ullah, and M.~Guizani, ``Vision-language models for edge networks: A comprehensive survey,'' \emph{arXiv preprint arXiv:2502.07855}, 2025. [Online]. Available: \url{https://arxiv.org/html/2502.07855v1}
\BIBentrySTDinterwordspacing

\bibitem{turn0search13}
\BIBentryALTinterwordspacing
Y.~Jin, J.~Li \emph{et~al.}, ``Efficient multimodal large language models: A survey,'' \emph{arXiv preprint arXiv:2405.10739}, 2024. [Online]. Available: \url{https://arxiv.org/html/2405.10739v1}
\BIBentrySTDinterwordspacing

\bibitem{turn0search2}
J.~Luo, C.~Chen, and S.~Wu, ``Federated prompt learning for vision-language models,'' \emph{arXiv preprint arXiv:2410.10114}, 2024.

\bibitem{turn0academia11}
------, ``Mixture of experts made personalized: Federated prompt learning for vision-language models,'' \emph{arXiv preprint arXiv:2410.10114}, 2024.

\bibitem{turn0academia8}
D.~P. Nguyen, J.~P. Munoz, and A.~Jannesari, ``Flora: Enhancing vision-language models with parameter-efficient federated learning,'' \emph{arXiv preprint arXiv:2404.15182}, 2024.

\bibitem{turn0search0}
B.~Pan, W.~Huang, and Y.~Shi, ``Federated learning from vision-language foundation models,'' \emph{arXiv preprint arXiv:2409.19610}, 2024.

\bibitem{turn0academia10}
G.~Sun, M.~Mendieta, A.~Dutta, X.~Li, and C.~Chen, ``Towards multi-modal transformers in federated learning,'' \emph{arXiv preprint arXiv:2404.12467}, 2024.

\bibitem{gopalkrishnan2024emvlm4ad}
\BIBentryALTinterwordspacing
A.~Gopalkrishnan, R.~Greer, and M.~Trivedi, ``Multi-frame, lightweight \& efficient vision-language models for question answering in autonomous driving,'' 2024. [Online]. Available: \url{https://arxiv.org/abs/2403.19838}
\BIBentrySTDinterwordspacing

\bibitem{cheng2024litevila}
Y.~Cheng, M.-H. Chen, and S.-H. Lai, ``Litevila: A lightweight vision-language model for scene understanding in autonomous driving,'' \emph{ECCV 2024 Workshop W-CODA}, 2024.

\bibitem{curto2023uavsceneunderstanding}
J.~de~Curtò, I.~de~Zarzà, and C.~Calafate, ``Semantic scene understanding with large language models on unmanned aerial vehicles,'' \emph{Drones}, vol.~7, p. 114, 02 2023.

\bibitem{turn0search12}
\BIBentryALTinterwordspacing
Y.~Liu, X.~Wang \emph{et~al.}, ``Edge-assisted object segmentation using multimodal feature fusion,'' \emph{ACM Transactions on Sensor Networks}, vol.~19, no.~2, p.~22, 2023. [Online]. Available: \url{https://dl.acm.org/doi/10.1145/3612922}
\BIBentrySTDinterwordspacing

\bibitem{turn0search14}
\BIBentryALTinterwordspacing
H.~Alikhani, A.~Kanduri \emph{et~al.}, ``Dynafuse: Dynamic fusion for resource efficient multi-modal machine learning inference,'' \emph{arXiv preprint arXiv:2306.15333}, 2023. [Online]. Available: \url{https://www.researchgate.net/publication/374181098_DynaFuse_Dynamic_Fusion_for_Resource_Efficient_Multi-Modal_Machine_Learning_Inference}
\BIBentrySTDinterwordspacing

\bibitem{turn0search10}
\BIBentryALTinterwordspacing
Y.~Zhang, J.~Li, Q.~Wang \emph{et~al.}, ``Few-shot learning with multimodal fusion for efficient cloud–edge collaborative mmwave beam selection,'' \emph{Electronics}, vol.~14, no.~4, p. 804, 2025. [Online]. Available: \url{https://www.mdpi.com/2079-9292/14/4/804}
\BIBentrySTDinterwordspacing

\bibitem{zhu2024crossmodality}
T.~Zhu \emph{et~al.}, ``Unraveling cross-modality knowledge conflicts in large vision-language models,'' \emph{arXiv preprint arXiv:2410.03659}, 2024.

\bibitem{cheng2024visualprompt}
Y.~Cheng, M.-H. Chen, and S.-H. Lai, ``Visual prompt multi-modal tracking,'' in \emph{Proceedings of the Conference on Computer Vision and Pattern Recognition (CVPR)}, 2024.

\bibitem{tian2023cmt}
S.~Tian \emph{et~al.}, ``Cross-modality fusion for depth prediction via rgb and thermal data,'' \emph{IEEE Xplore}, 2023.

\bibitem{usynin2022beyond}
D.~Usynin, D.~Rueckert, and G.~Kaissis, ``Beyond gradients: Exploiting adversarial priors in model inversion attacks,'' \emph{ArXiv}, 2022.

\bibitem{zhou2023boosting}
S.~Zhou, T.~Zhu, D.~Ye, X.~Yu, and W.~Zhou, ``Boosting model inversion attacks with adversarial examples,'' \emph{ArXiv}, 2023.

\bibitem{zhang2022towards}
J.~Zhang, Q.~Yi, and J.~Sang, ``Towards adversarial attack on vision-language pre-training models,'' in \emph{Proceedings of the 30th ACM International Conference on Multimedia}, 2022.

\bibitem{wen2021defending}
J.~Wen, S.~Yiu, and L.~Hui, ``Defending against model inversion attack by adversarial examples,'' in \emph{2021 IEEE International Conference on Cyber Security and Resilience (CSR)}, 2021, pp. 551--556.

\bibitem{rathore2023formal}
\BIBentryALTinterwordspacing
A.~Rathore, M.~Blanton, M.~Gaboardi, and L.~Ziarek, ``A formal model for secure multiparty computation,'' 2023. [Online]. Available: \url{https://arxiv.org/abs/2306.00308}
\BIBentrySTDinterwordspacing

\bibitem{lu2023setlevel}
D.~Lu, Z.~Wang, T.~Wang, W.~Guan, H.~Gao, and F.~Zheng, ``Set-level guidance attack: Boosting adversarial transferability of vision-language pre-training models,'' \emph{ArXiv}, 2023.

\bibitem{park2022privacy}
J.~Park and H.-K. Lim, ``Privacy-preserving federated learning using homomorphic encryption,'' \emph{Applied Sciences}, 2022.

\bibitem{park2022federated}
------, ``Privacy-preserving federated learning using homomorphic encryption,'' \emph{Applied Sciences}, vol.~12, p. 734, 2022.

\bibitem{fang2021privacy}
H.~Fang and Q.~Qian, ``Privacy preserving machine learning with homomorphic encryption and federated learning,'' \emph{Future Internet}, vol.~13, p.~94, 2021.

\bibitem{shen2023privacy}
C.~Shen and W.~Zhang, ``Privacy enhanced federated learning via privacy masks and additive homomorphic encryption,'' in \emph{2023 International Conference on Networking and Network Applications (NaNA)}, 2023, pp. 471--478.

\bibitem{hussien2023secure}
N.~Hussien, S.~A. Salman, and M.~Aljanabi, ``Secure federated learning with a homomorphic encryption model,'' \emph{International Journal Papier Advance and Scientific Review}, 2023.

\bibitem{park2022homomorphic}
J.~Park, N.~Y. Yu, and H.~Lim, ``Privacy-preserving federated learning using homomorphic encryption with different encryption keys,'' in \emph{2022 13th International Conference on Information and Communication Technology Convergence (ICTC)}, 2022, pp. 1869--1871.

\bibitem{ma2021privacy}
J.~Ma, S.-A. Naas, S.~Sigg, and X.~Lyu, ``Privacy‐preserving federated learning based on multi‐key homomorphic encryption,'' \emph{International Journal of Intelligent Systems}, vol.~37, pp. 5880--5901, 2021.

\bibitem{du2022signcryption}
W.~Du, M.~Li, Y.~Han, X.~A. Wang, and Z.~Wei, ``A homomorphic signcryption-based privacy preserving federated learning framework for iots,'' \emph{Security and Communication Networks}, 2022.

\bibitem{sébert2023combining}
A.~G. Sébert, M.~Checri, O.~Stan, R.~Sirdey, and C.~Gouy-Pailler, ``Combining homomorphic encryption and differential privacy in federated learning,'' in \emph{2023 20th Annual International Conference on Privacy, Security and Trust (PST)}, 2023, pp. 1--7.

\bibitem{naveen2022memory}
S.~Naveen and M.~R. Kounte, ``Memory optimization at edge for distributed convolution neural network,'' \emph{Transactions on Emerging Telecommunications Technologies}, 2022.

\bibitem{shaowang2021declarative}
T.~Shaowang, N.~Jain, D.~Matthews, and S.~Krishnan, ``Declarative data serving: The future of machine learning inference on the edge,'' \emph{Proc. VLDB Endow.}, vol.~14, pp. 2555--2562, 2021.

\bibitem{fu2023energy}
Z.~Fu, A.~Avaliani, and M.~Donato, ``Energy-efficient task adaptation for nlp edge inference leveraging heterogeneous memory architectures,'' \emph{ArXiv}, 2023.

\bibitem{hu2021pipeline}
Y.~Hu, C.~Imes, X.~Zhao, S.~Kundu, P.~Beerel, S.~Crago, and J.~Walters, ``Pipeline parallelism for inference on heterogeneous edge computing,'' \emph{ArXiv}, 2021.

\bibitem{xu2023devit}
G.~Xu, Z.~Hao, Y.~Luo, H.~Hu, J.~An, and S.~Mao, ``Devit: Decomposing vision transformers for collaborative inference in edge devices,'' \emph{ArXiv}, 2023.

\bibitem{dutta2023search}
O.~Dutta, T.~Kanvar, and S.~Agarwal, ``Search-time efficient device constraints-aware neural architecture search,'' \emph{ArXiv}, 2023.

\end{thebibliography}

\begin{IEEEbiography}
[{\includegraphics[width=1in,height=1.25in,clip,keepaspectratio]{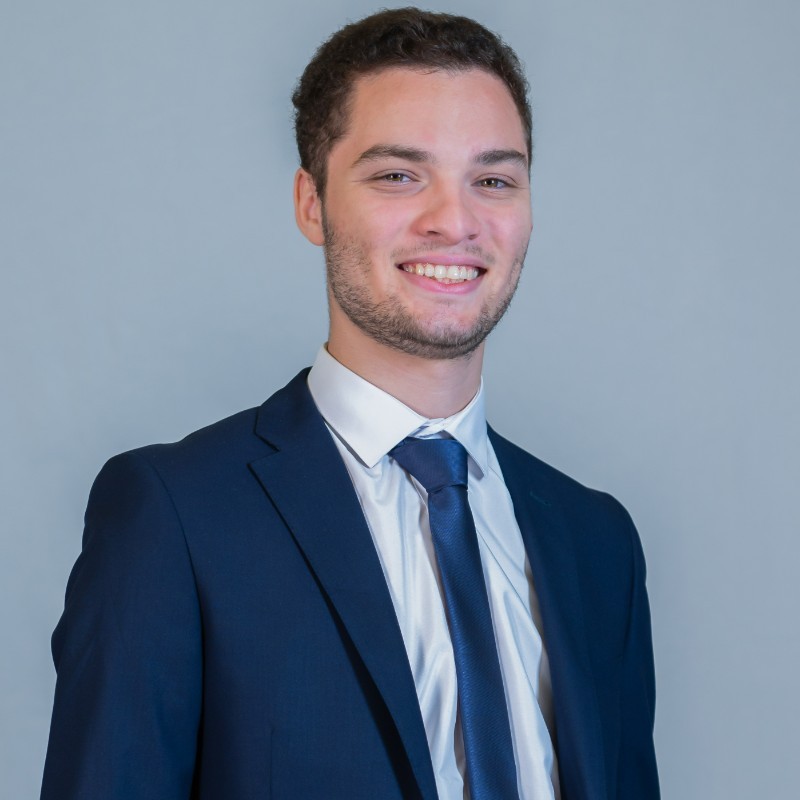}}]{Ahmed Sharshar} is currently pursuing a PhD in Computer Vision at Mohamed bin Zayed University of Artificial Intelligence (MBZUAI) in Abu Dhabi, UAE. He previously received his Master of Science degree in Computer Vision from MBZUAI. He obtained his Bachelor of Engineering degree in Computer Engineering from the Egypt-Japan University of Science and Technology (E-JUST), Egypt.
His research primarily focuses on developing lightweight models and expanding their applications across various domains, such as natural language processing, computer vision, and human-computer interaction. Specifically, he aims to make these models more efficient and accessible, ensuring broader usability and practical deployment.
\end{IEEEbiography}

\begin{IEEEbiography}[{\includegraphics[width=1in,height=1.25in,clip,keepaspectratio]{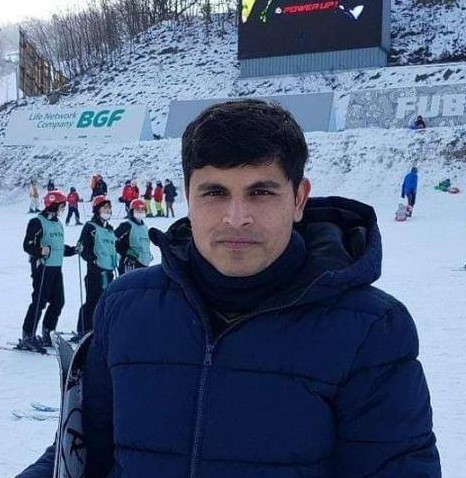}}]{Latif U. Khan} received his Ph.D. degree in Computer Engineering at Kyung Hee University (KHU), South Korea. Prior to that, He received his MS (Electrical Engineering) degree with distinction from University of Engineering and Technology, Peshawar, Pakistan in 2017. He is the recipient of KHU Best thesis award. He is author of two books: (a) Network Slicing for $5$G and Beyond and (b) Federated Learning for Wireless Networks. He has reviewed over 200 times for the top ISI- Indexed journals and conferences. He has authored many most popular articles in the leading journals (i.e., IEEE Communications Surveys and Tutorials) and magazines (IEEE Communication Magazine, IEEE Network, and IEEE Wireless Communications Magazine). His research interests include analytical techniques of optimization and game theory to edge computing, end-to-end network slicing, wireless federated learning, and digital twins. 
\end{IEEEbiography}

\begin{IEEEbiography}[{\includegraphics[width=1in,height=1.25in,clip,keepaspectratio]{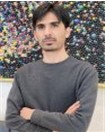}}]{Waseem Ullah} (Student Member, IEEE) received his M.S. degree in Computer Science from Islamia College Peshawar, Pakistan, in 2019. He completed his Ph.D. program at Sejong University, Seoul, South Korea, with the Intelligent Media Laboratory (IM Lab). Currently, he is serving as a Postdoctoral Fellow at the Mohamed bin Zayed University of Artificial Intelligence (MBZUAI), UAE. His research interests include computer vision techniques, anomaly detection, bioinformatics, pattern recognition, deep learning, image processing, video analysis, and medical image analysis. 
\end{IEEEbiography}

\begin{IEEEbiography}[{\includegraphics[width=1.5in,height=1.25in,clip,keepaspectratio]{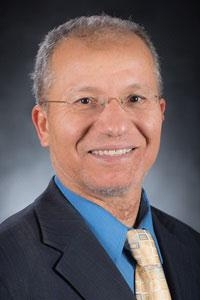}}]{Mohsen Guizani} (S’85, M'89, SM'99, F’09) received his B.S. (with distinction) and M.S. degrees in electrical engineering, and M.S. and Ph.D. degrees in computer engineering from Syracuse University, New York, in 1984, 1986, 1987, and 1990, respectively. He is currently a Professor with the Machine Learning Department, Mohamed Bin Zayed University of Artificial Intelligence (MBZUAI), Abu Dhabi, UAE. Previously, he served in different academic and administrative positions at the University of Idaho, Western Michigan University, the University of West Florida, the University of Missouri-Kansas City, the University of Colorado-Boulder, and Syracuse University. His research interests include wireless communications and mobile computing, computer networks, mobile cloud computing, security, and smart grid. 
\end{IEEEbiography}

\end{document}